\pdfoutput=1
\documentclass[11pt]{article}
\usepackage[]{acl}
\usepackage{times}
\usepackage{latexsym}
\usepackage{graphicx}
\usepackage{tabularx}
\usepackage{booktabs}
\usepackage{multirow}
\usepackage{adjustbox}
\usepackage{amsmath}
\usepackage{multicol}
\usepackage{titlesec}
\usepackage{xcolor}         
\usepackage{colortbl}         
\usepackage{amssymb} 
\usepackage{breqn}
\usepackage{enumitem}
\usepackage{pgfplots}
\usepackage{pgfplotstable}
\usepackage{makecell}
\usepackage{pgfplots}
\pgfplotsset{compat=1.17}
\usepackage[T1]{fontenc}
\usepackage[utf8]{inputenc}
\usepackage{microtype}
\usepackage{inconsolata}
\usepackage{cleveref}

\newcommand{\xhdr}[1]{\vspace{2mm} \noindent {\bf #1.}}

\usepackage{xspace}
\newcommand{\shortpapername}{EditInspector\xspace}
\newcommand{\papername}{\shortpapername: A Benchmark for Evaluation of Text-Guided Image Edits\xspace}

\makeatletter
\newcommand\sparagraph[1]{\@startsection{paragraph}{4}{\z@}%
  {#1}
  {-1em}
  {\normalfont\normalsize\bfseries\noindent}*}
\makeatother

\newcommand{\com}[1]{}
\newcommand{\resolved}[1]{}

\title{\papername}
\author{
  \textbf{Ron Yosef\textsuperscript{1}},
  \textbf{Moran Yanuka\textsuperscript{2}},
  \textbf{Yonatan Bitton\textsuperscript{3}},
  \textbf{Dani Lischinski\textsuperscript{1}}
\\
  \textsuperscript{1}The Hebrew University of Jerusalem,  
  \textsuperscript{2}Tel Aviv University,  
  \textsuperscript{3}Google Research,
\\
}

\date{Oct 2024}

\begin{document}

\maketitle

\begin{abstract}
Text-guided image editing, fueled by recent advancements in generative AI, is becoming increasingly widespread. This trend highlights the need for a comprehensive framework to verify text-guided edits and assess their quality. To address this need, we introduce \shortpapername, a novel benchmark for evaluation of text-guided image edits, based on human annotations collected using an extensive template for edit verification\footnote{https://editinspector.github.io/}. We leverage \shortpapername to evaluate the performance of state-of-the-art (SoTA) vision and language models in assessing edits across various dimensions, including accuracy, artifact detection, visual quality, seamless integration with the image scene, adherence to common sense, and the ability to describe edit-induced changes. Our findings indicate that current models struggle to evaluate edits comprehensively and frequently hallucinate when describing the changes. To address these challenges, we propose two novel methods that outperform SoTA models in both artifact detection and difference caption generation.
\end{abstract}

\section{Introduction}
\label{sec:introduction}
The ability to create and modify images is vital in fields such as social media, marketing, and graphic design. Recent advancements in generative AI have greatly democratized this ability. In particular, natural language enables high-quality, customized visual content creation with minimal effort.

Text-guided editing models require a source image and instruction \cite{kawar2023imagic, zhang2022sine, brooks2023instructpix2pix, wu2023selfcorrecting, zhang2024hive}, sometimes allowing multi-turn editing \cite{sheynin2023emu, he2024llms, wu2023visual, cui2023chatedit}. 
For more precise spatial control a user might provide the source image, a mask, and a text prompt specifying changes for the masked area \cite{Avrahami_2022, nichol2022glide, couairon2022diffedit, Editbench, Magicbrush}. 
Extensive human evaluations showed that mask-based text-guided editing produces superior results compared to mask-free editing \cite{Editbench, Magicbrush}.

\begin{figure*}[tp]
    \centering
    \includegraphics[width=1\textwidth]{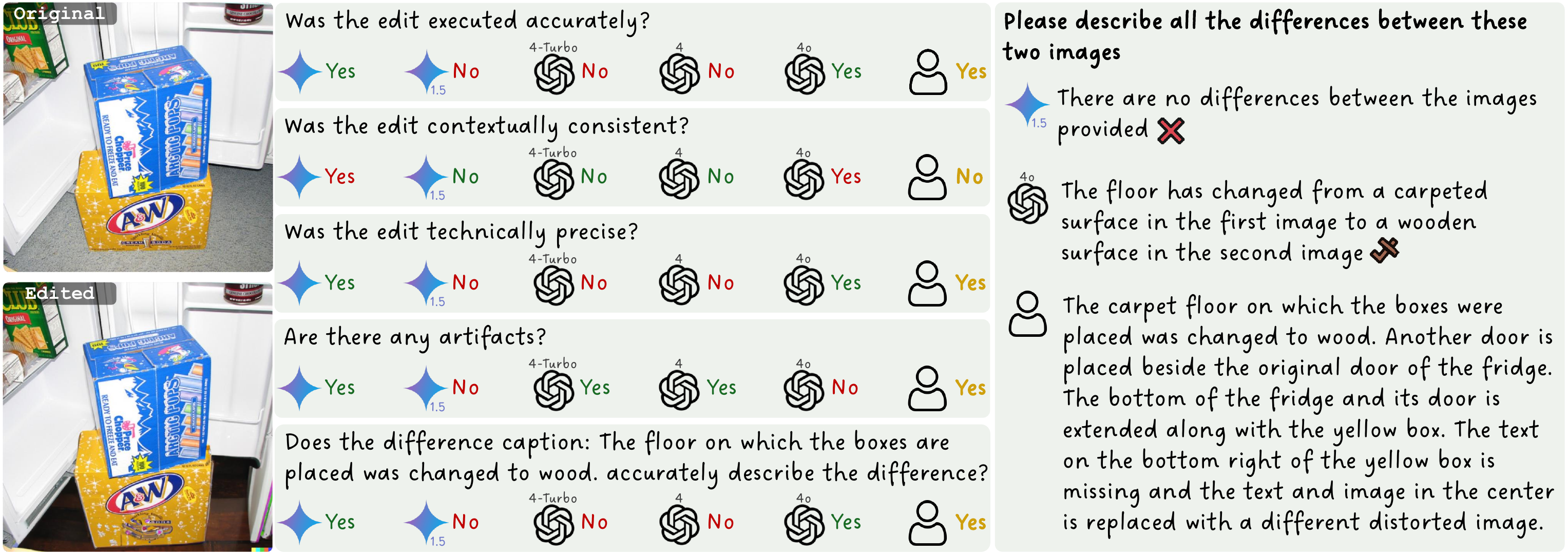}
    \caption{The assessments for the edit ``Let the floor be made of wood'' vary across different models, with 2--3 models answering each question correctly. Gemini 1.5 failed to detect any differences between the images, while GPT-4o successfully identified only the main difference. See Appendix~\ref{sec:appendx_taks_prompts} for full-size prompts.}
    \label{fig:models_results_example}
\end{figure*}

\begin{figure*}[tp]
    \centering
    \includegraphics[width=0.9\textwidth]{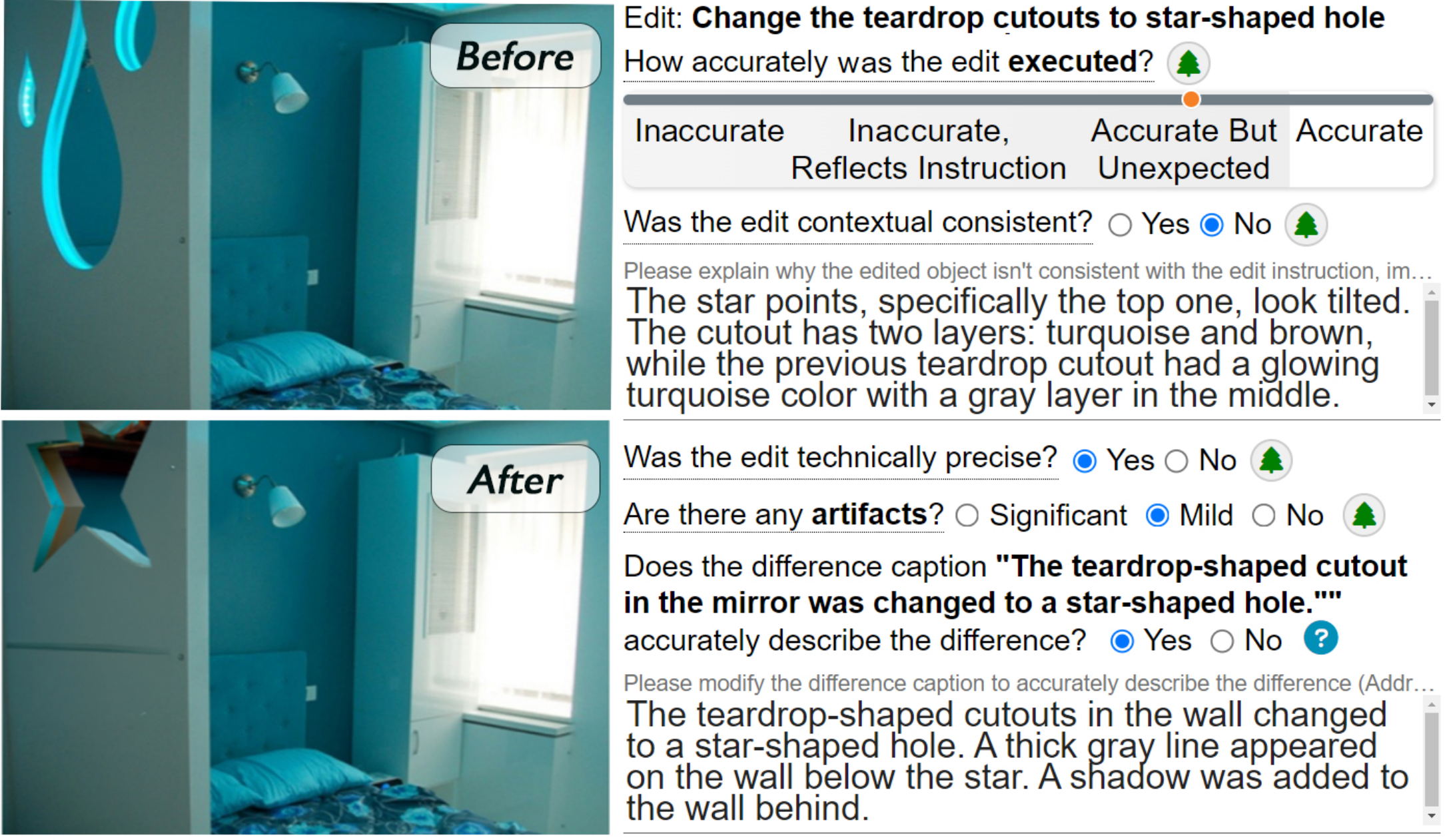}
    \caption{This is an example of our annotation user interface. The edit appears to be accurately executed but includes unexpected elements, such as differences in the door layers and a tilted star edge. There are mild artifacts, including a shadow behind the wall and a thick gray line beneath the star cutout. Clicking the tree icons opens decision trees that help annotators follow the evaluation guidelines (See \Cref{sec:appendx_annotation_ui}).}
    \label{fig:user-interface}
\end{figure*}

Despite these advancements, evaluating the quality and accuracy of edits remains challenging, as demonstrated in \Cref{fig:models_results_example}. Current methods often focus on whether the edited object matches the requested attributes \cite{Editbench} or use ranking scores for accuracy \cite{Magicbrush}. However, they overlook pain points such as unintended artifacts, misalignment with user expectation, visual quality, and adherence to common sense. For example, in \Cref{fig:user-interface}, the edit changes teardrops to stars as instructed, but unintentionally adds a line and alters the wall’s appearance.

To address these challenges, we propose \shortpapername, a comprehensive benchmark for \emph{assessing evaluators of text-guided image edits} (\Cref{sec:dataset}). \shortpapername examines edits across five dimensions: (1) whether the edit accurately follows the instructions and aligns with user expectations; (2) introduction of unintended artifacts; (3) technical quality (low resolution, blur, etc.); (4) the accuracy of a description of the main difference; and (5) the accuracy of a detailed listing of the differences between the original and the edited images.

We begin by creating a human evaluation framework, shown in \Cref{fig:user-interface}, that assesses edits based on the dimensions outlined above (\Cref{sec:human_evaluation_framework}). Using this framework, we collected human annotations as edit inspectors through crowdsourcing, evaluating 783 edits from the MagicBrush \cite{Magicbrush} test set of 1,053 edits, to introduce the \shortpapername benchmark (\Cref{sec:human_annotation}).

We then evaluate state-of-the-art vision and language models (VLMs) as edit inspectors on the EditInspector benchmark, comparing their performance with human annotations, as shown in Figure~\ref{fig:models_results_example}. The results show that all models perform poorly across all tasks, with accuracy hovering around random chance (Section~\ref{sec:basic_edit_verification_results}). Gemini-1.5 \cite{geminiProVision1.5} emerged as the top performer for the edit inspector questions, achieving 70.3\% accuracy in the edit accuracy question. We evaluate models' ability to generate a summary of the main change and a detailed list of all differences as an upper-bound test of edit accuracy, artifact detection, and visual quality. In this task, GPT-4o achieved 39\% accuracy in describing the main difference but detected only 12\% of all differences, with only 40\% aligning with human annotations, highlighting significant hallucinations. (Section \ref{sec:difference_caption_generation_results}).

We tackle the challenges of artifact detection and difference caption generation with two methods. First, we developed a zero-shot pipeline using Gemini as the visual backbone to generate instruction-grounded difference captions and metadata (\Cref{sec:difference_caption_pipeline}). The pipeline analyzes image captions at three zoom levels around the edit area and outputs a difference caption, achieving 75\% accuracy in describing the main difference, compared to 39\% by the best SoTA model. Second, we introduced a novel artifact detection method that achieves 64\% accuracy by analyzing object segmentation probabilities around the edited area (\Cref{sec:artifacts_detection}).

Finally, we introduce an end-to-end fine-tuned model that rivals much larger models, delivering competitive SoTA performance while reducing computational costs (\Cref{sec:supervision}).
To train our model we use two augmentation methods to generate 31,059 training instances. The first method creates negative examples with objects closely resembling the original (\Cref{sec:negative_edit_augmenation}), and the second reverses the edit direction, e.g., by changing an ``Add'' edit to a ``Remove'' edit (\Cref{sec:reverse_edit_augmenation}). 

In summary, our main contributions are: (1) A comprehensive framework for image edit evaluation, and the \shortpapername benchmark, which we release for future work and future model assessment; (2) A thorough evaluation of SoTA VLMs as edit inspectors, showing that, across all aspects, none can effectively assess edits; (3) Two new methods outperforming SoTA models for artifact detection and difference caption generation; and, (4) An end-to-end fine-tuned model that rivals much larger models in performance.
\section{\shortpapername Dataset}
\label{sec:dataset}
Our goal is to develop a dataset and framework for image editing verification that offers a comprehensive evaluation of edits, addressing overlooked pain points like unintended artifacts, instruction inconsistencies, scene misalignment, and technical flaws.
To achieve this, we introduced the human evaluation framework in \Cref{sec:human_evaluation_framework} and annotated 783 MagicBrush edits using it to create our benchmark in \Cref{sec:human_annotation}. MagicBrush is a manually annotated dataset for instruction-guided, mask-provided image editing. The statistics and analysis of our benchmark are presented in \Cref{sec:dataset_statistics_analysis}.

\subsection{Human Evaluation Framework}
\label{sec:human_evaluation_framework}
Our motive was to develop a comprehensive framework that evaluates multiple aspects of image editing. We tested and refined templates and questions using internal and crowdsourced feedback, resulting in the framework shown in \Cref{fig:user-interface}.

The evaluation begins with \emph{Accuracy Level}, where annotators assess whether the edit follows the instruction and meets user expectations. If it fully follows the instruction, annotators select \emph{Accurate} or \emph{Accurate, But Unexpected} if it deviates from expectations. For partial adherence to the instruction, they select \emph{Inaccurate, Reflects Instruction}, and for no adherence, \emph{Inaccurate}.

For any selection other than \emph{Accurate}, annotators are asked to explain under \emph{Contextual Consistency} how the edit failed to meet expectations or align with the instruction, image scene, or common sense. Under the \emph{Technical Precision} question annotators comment on pixel-level details like resolution, blurriness, and smoothness. 

For example, in Figure~\ref{fig:user-interface} a teardrop cutout was changed to a star-shaped hole, but all annotators marked it as ``\emph{Accurate, But Unexpected}'' due to the tilted star edge and the unexpected material appearance, as seen in \emph{Contextual Consistency} feedback.

Next, the \emph{Artifacts} evaluation involves annotators identifying any unintended distortions or anomalies in the edit. Artifacts are classified into two levels: \emph{Significant} or \emph{Mild}, based on their severity. In the example in \Cref{fig:user-interface}, two \emph{Mild} artifacts are present: an unintended shadow and an extra line beneath the star-shaped hole.

Finally, to collect a difference caption that describes all differences between the original and edited images as an upper-bound evaluation, we start with an automatically generated caption that describes the main difference (\Cref{sec:difference_caption_pipeline}). Humans then review it, either accepting or correcting it and expand it to include additional differences if artifacts are present, as shown in \Cref{fig:user-interface}.

\subsection{Human Annotation}
\label{sec:human_annotation}
We employed Amazon Mechanical Turk (AMT) to evaluate image edits using human annotators, as shown in \Cref{fig:user-interface}, with three annotations per edit. Quality annotators were selected through a paid qualification test, and multiple steps were taken to ensure the instructions were clear and accessible in the UI (See \Cref{sec:additional_annotation_information,sec:appendx_annotation_examples}).

\subsection{Human Evaluation Analysis}
\label{sec:dataset_statistics_analysis}
Majority vote label distribution is presented in \Cref{tab:task_annotation_stats}. Despite the task's subjectivity, agreement averaged 80\% to 86\%, compared to random chance of 25\% for Accuracy and 33\% for Artifacts. Majority agreement hit 95\% for Accuracy and 97\% for Artifacts. Full agreement among all annotators was achieved for 42\% to 57\% of edits.
In 85\% of examples, the edit reflected the instruction (``Accurate'' or ``Accurate, But Unexpected''), while 38\% of edits contained significant artifacts.


The edit types were distributed as follows: Add 35.8\%, Change Attribute 21.6\%, Remove 7.3\%, and Replace 31.3\%, based on the image caption pipeline metadata in \Cref{sec:difference_caption_pipeline}.
\begin{table}[tp]
    \centering
    \small
    \begin{tabular}{lcc}
        \toprule
        Category                          & \multicolumn{2}{c}{Statistics (\%)} \\
        \midrule
        \makecell{Accuracy \\ Level}       & \makecell[l]{Accurate: 8\% \\ Accurate \\ Unexpected: 77\%} & \makecell[l]{Inaccurate: 6\% \\ Inaccurate \\ Reflects: 4\%} \\
        \midrule
        \makecell{Artifacts \\ Level}      & \makecell[l]{Significant: 38\%} & \makecell[l]{Mild: 57\% \\ No Artifact: 2\%} \\
        \midrule
        \makecell{Technical \\ Precision}  & \makecell[l]{Yes: 31\%} & \makecell[l]{No: 69\%} \\
        \midrule
        \makecell{Visual \\ Consistency}   & \makecell[l]{Yes: 18\%} & \makecell[l]{No: 82\%} \\
        \midrule
        \makecell{Diff Caption\\ Accuracy} & \makecell[l]{Yes: 60\%} & \makecell[l]{No: 40\%} \\
        \bottomrule
    \end{tabular}
    \caption{Distribution of majority vote labels across categories. In 85\% of examples, the edit reflected the instruction (``Accurate'' or ``Accurate, But Unexpected''), while 38\% of edits contained significant artifacts. }
    \label{tab:task_annotation_stats}
\end{table}

\begin{figure}[tp]
    \small
    \centering
    \begin{tikzpicture}
        \begin{axis}[
            ybar,
            bar width=10pt, 
            width=0.48\textwidth, 
            height=0.4\textwidth, 
            symbolic x coords={Shape/Proportion, Blur/Fuzziness, Texture, Lighting/Brightness, Color, Unreal/Artificial look, Placement, Missing/Extra Objects, Edges, Resolution},
            xtick=data,
            x tick label style={rotate=45, anchor=north east, font=\footnotesize}, 
            ymin=0,
            ymax=100,
            ytick={0,10,20,30,40,50,60,70,80,90,100}, 
            y tick label style={/pgf/number format/.cd, fixed, precision=0, /tikz/.cd, font=\scriptsize}, 
            yticklabel={\pgfmathprintnumber{\tick}\%}, 
            nodes near coords,
            every node near coord/.append style={font=\scriptsize}, 
            grid=major, 
            major grid style={line width=0.2pt,draw=gray!50}, 
            enlarge x limits=0.2, 
        ]
        
        \addplot coordinates {
            (Shape/Proportion, 75) 
            (Blur/Fuzziness, 73) 
            (Texture, 67) 
            (Lighting/Brightness, 50) 
            (Color, 46) 
            (Unreal/Artificial look, 47) 
            (Placement, 16) 
            (Missing/Extra Objects, 7) 
            (Edges, 34) 
            (Resolution, 83)
        };
        
        \end{axis}
    \end{tikzpicture}
    \caption{Frequency of issues identified by human annotators in the Contextual Consistency and Technical Precision textual feedback. Shape/Proportion concerns being particularly prominent.}
    \label{tab:feedback_issues}
\end{figure}
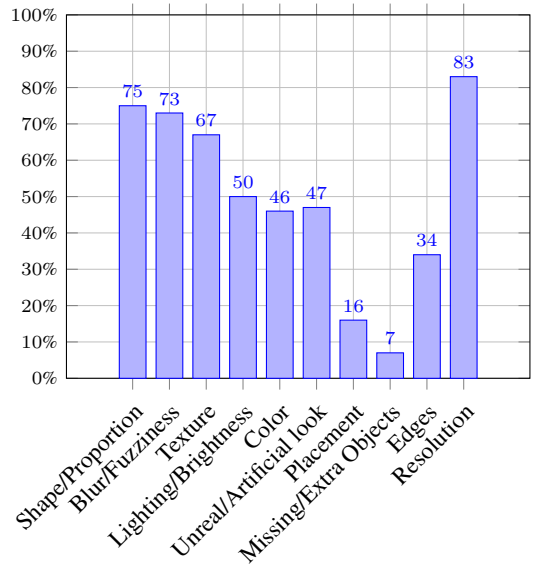

\Cref{tab:feedback_issues} shows the percentage of issues reported by annotators in the Contextual Consistency and Technical Precision feedback, with resolution and shape/proportion concerns being particularly prominent. See \Cref{sec:textual_feedback_categories} for a full overview.

\section{Auto-Evaluation}
\label{sec:expirments}
Using the \shortpapername benchmark, we evaluate the ability of SoTA VLMs to serve as edit inspectors. The evaluation consists of two components: the first assesses the models’ ability to verify edit accuracy and alignment with user expectations, while the second serves as an upper-bound test, examining their ability to generate captions that describe the main differences and all differences, including unintended artifacts (Section \ref{sec:difference_caption_generation_results}).

\subsection{Models}
We evaluate GPT-4, GPT-4o, GPT-4-turbo \cite{openai2024gpt4technicalreport}, Gemini-Pro-Vision \cite{geminiProVision}, Gemini-Pro-1.5 \cite{geminiProVision1.5}, Qwen2.5-VL \cite{bai2025qwen25vltechnicalreport} and InternVL3 \cite{zhu2025internvl3exploringadvancedtraining} on all tasks using their latest versions as of August 2024 (Section \ref{sec:model_versions}). We prioritized prompts that best conveyed user instructions and improved overall performance (See \Cref{sec:appendx_taks_prompts}). 

\subsection{Auto-Evaluation Setup}
\label{sec:metrics}
\xhdr{Edit Inspector Questions} Preliminary experiments revealed that models struggled to handle multiple categories, especially in detecting mild artifacts. 
To enhance clarity and relevance, we simplified the categorization by replacing multiple-choice questions with binary questions. For the accuracy question, both ``Accurate'' and ``Accurate But Unexpected'' were grouped under ``Accurate,'' while in the artifacts question, only ``Significant Artifacts'' were counted as artifacts.

\xhdr{Difference Caption Generation}
Traditional caption metrics (BLEU, METEOR, ROUGE, CIDEr) rely on N-gram overlaps but fail to distinguish edited objects, penalize stylistic variations, ignore edit sequences, and miss semantic misalignments. As shown in Table~\ref{tab:metric_comparison}, these limitations lead to misleadingly high scores for incorrect captions. Section~\ref{sec:common_metrics_for_caption_comperasion_overview} provides further examples and analysis.

To address these limitations, we propose two novel evaluation metrics tailored for \textbf{all differences caption} comparisons: Model Precision (MP) and Hallucination Rate (HR). MP is the percentage of human-annotated differences matching model-detected ones, while HR is the percentage of model-detected differences that do not correspond to any human-annotated differences.

We calculate these metrics by generating Difference Triplets (DTs) with the source object, target object, and action type for each change in the model and human captions. The two resulting sets of DTs are then used to compute MP and HR. A match between two DTs is determined if the edit action types are identical, and the source and target objects are similar, as evaluated by GPT-4o. The similarity check between source and target objects is relaxed, allowing matches for objects with different attributes. A stricter check would have caused models to fail completely.

In addition, we introduced MP\textsubscript{soft} and HR\textsubscript{soft}, which count DT matches also in case of a reversed source and target object match, offering a more comprehensive analysis of model performance. See Section \ref{sec:metics_mathematical_explenations} for mathematical formulations of the metrics, and Section \ref{sec:metics_example} for an intuitive example.

We evaluate the model's \textbf{main difference caption} by comparing it to the main difference extracted from the human-provided difference caption, which describes all of the edit's differences. GPT-4 is used to assess whether the main model-identified difference matches the human one. Extracting the main difference is not complex, as the main change is typically mentioned first.

\begin{table}[tp]
    \small
    \setlength{\tabcolsep}{2pt}
    \centering
    \begin{tabular}{p{6.2cm}@{\hskip 5pt}c}
        \hline
        Example & Metrics \\
        \midrule
        \begin{tabular}[x]{@{}p{6.2cm}@{}}\textbf{Ground Truth Caption:} The main difference is the first image has a blue vase, and the second image has a brown vase. \\ \textbf{Generated Caption:} The main difference is the first image has a squirrel, and the second image does not. \end{tabular}
        & \makecell{\textbf{MP}: 0 \\ BL: 0.68 \\ RO: 0.81 \\ ME: 0.78} \\
        \midrule
        \begin{tabular}[x]{@{}p{6.2cm}@{}}\textbf{Ground Truth Caption:} A brown squirrel was added to the image. \\ \textbf{Generated Caption:} The difference between the two images is that the first image has a blue vase. The second image has a blue vase and a squirrel next to it. \end{tabular}
        & \makecell{\textbf{MP}: 1\\ BL: 0.55 \\ RO: 0.60 \\ ME: 0.57}\\
       \midrule
        \begin{tabular}[x]{@{}p{6.2cm}@{}}\textbf{Ground Truth Caption:} In the first image, the tree was removed, and new flowerbed was added. \\ \textbf{Generated Caption:} In the first image, the flowerbed was removed, and new tree was added. \end{tabular}
        & \makecell{\textbf{MP}: 0 \\ BL: 0.73 \\ RO: 0.79 \\ ME: 0.76} \\
        \hline
    \end{tabular}
    \caption{Comparison of traditional linguistic metrics (BLEU, ROUGE, METEOR) against our proposed evaluation metric (MP). The first example shows high scores despite missing the edited object. The second penalizes correct but longer captions. The third fails to detect reversed edits, while our metric captures these issues.}
    \label{tab:metric_comparison}
\end{table}

\subsection{Auto-Evaluation Results}
\label{sec:evaluation_results}
\begin{table*}[tp]
    \centering
    \small
    \begin{tabular}{lcccccccccc}
        \toprule
                                 & Gemini            & Gemini-1.5       & GPT-4         & GPT-4o          & \makecell{GPT-4 \\ Turbo}  & \makecell{Qwen2.5 \\ VL}  & \makecell{InternVL3} & LLaVA  &  \makecell{LLaVA \\ (Supervised)}\\
        \midrule
        \multicolumn{6}{c}{\textbf{Edit Inspectors Questions}} \\
        \midrule
        \makecell[l]{Accuracy}                    & 49.9\%         & \textbf{70.3\%}  & 67.3\%        & 67.8\%          & 66.9\%         & 67.7\%     & 70.2\%      & 58.9\%  & 67.2\% \\
        \makecell[l]{Contextual  \\  Consistency}      & 50.4\%         & 51.1\%           & 50.4\%        & \textbf{55.7\%}  & 48.2\%   & 49.5\%     & 49.2\%      & 52.0\%  & - \\
        \makecell[l]{Technical  \\  Precision}         & 50.1\%         & 46.3\%  & 53.7\%        & \textbf{55\%}           & 49.3\%     & 48.4\%     & 46.7\%      & 50.1\%  & - \\
        \makecell[l]{Artifacts}                   & 49.4\%         & 58.5\%           & 50.7\%        & \textbf{65.7\%} & 52.8\%         & 50\%       & 49.8\%      & 47.6\%  & 51.7\% \\
        \makecell[l]{Difference  \\ Caption Acc}  & 53.9\%         & \textbf{66.3\%}  & 63.9\%        & 64.3\%          & 64\%           & 58.2\%     & 58.2\%      & 50.0\%  & 54.5\% \\
        \midrule
        \multicolumn{6}{c}{\textbf{Differences Caption Generation}} \\
        \midrule
        \makecell[l]{Main  \\  Difference}            & 31\%            & 31\%           & 27\%        & \textbf{39\%} & 24\%      & 38\%          & 26\%     & 8\%     & 10\% \\
        MP                          & -              & 8\%              & 8\%           & \textbf{12\%}   & 8\%    & \textbf{12\%}          & 7\%      & -      & - \\
        MP\textsubscript{soft}      & -              & 9\%              & 10\%          & \textbf{14\%}   & 9\%    & \textbf{14\%}         & 10\%      & -      & - \\
        HR                          & -              & 67\%             & 78\%          & 60\%   & 75\%   & \textbf{58\%}          & 87\%      & -      & - \\
        HR\textsubscript{soft}      & -              & 65\%             & 75\%          & 56\%   & 72\%   & \textbf{52\%}          & 83\%      & -      & - \\
        \midrule
        Avg. Diff                   & -              & 1                & 2.5           & 1.8             & 1.5    & 1.5          & 3.1     & -      & - \\
        No Diffs                    & -              & 24\%             & 0.7\%         & 0.3\%           & 6\%    & 0.7\%          & 3.4\%      & -      & - \\
        \bottomrule
    \end{tabular}
    \caption{Combined performance on Edit Inspectors questions, and the Difference Caption Generation task. GPT-4o model demonstrates the best performance in Edit Inspectors questions, achieving the highest or third-highest scores across all questions. Qwen2.5-VL achieves the highest precision in predicting differences, with the lowest hallucination rate. Avg. Diff indicates the average number of differences detected per edit, while No Diffs represents the percentage of edits where no differences were predicted. Human annotators identified an average of 6 differences per edit. The main difference row reports the percentage of predicted main difference captions correctly describing the main difference. The LLaVA (Supervised) column presents the performance of the finetuned model; see Section 5.3 for further analysis.}
    \label{tab:benchmark_table}
\end{table*}

\paragraph{Edit Inspector Questions Results}
\label{sec:basic_edit_verification_results}
The results for the Yes/No questions are presented in \Cref{tab:benchmark_table}. GPT-4o achieved the highest score on most questions except `Edit Accuracy' and `Difference Caption Accuracy', where Gemini-1.5 scored the highest. Below, we summarize our main observations from these results.

\textbf{Struggling with Inaccurate Edits and Artifact Classification.} Detection of inaccurate edits was challenging, with most models correctly classifying only 0-30\%, except Qwen2.5-VL(39\%) and GPT-4o (47\%). All models mistakenly predicted edits as visually consistent, with precision scores between 0-22.3\%. Differentiating artifacts from non-artifacts was also challenging. While GPT-4o had the highest accuracy (65.7\%) it missed many artifacts with low recall (52.7\%). All models frequently misclassified non-artifacts (18-30\%), with Gemini misclassifying 72\%.

\textbf{Assessing the accuracy of inconsistent edits is challenging.} There is a strong conditional dependency between the edit accuracy and contextual consistency questions. A discrepancy up to 40\% was observed in the accuracy question when edits lacked contextual consistency. Conversely, models had difficulty with the contextual consistency question in accurate edits, with a 23\% drop in performance. This dependency was also present (up to 12\%) between the caption accuracy and contextual consistency questions.

\textbf{Remove edits are challenging for models.} While Gemini 1.5, GPT-4, GPT-4-turbo, and InternVL3 struggled with `Remove' edits, showing accuracy gaps up to 65\% in edit accuracy, GPT-4o excelled with 91\% accuracy, making it the only model to handle these edits well.

Alongside Yes/No questions, we assessed models' feedback on Contextual Consistency and Technical Precision, finding it misaligned with human feedback in most cases (see Appendix \ref{sec:textual_feedback}).

\paragraph{Difference Caption Generation Results}
\label{sec:difference_caption_generation_results}
\sparagraph{0.6ex}{Main Differences Captions:} 
Table \ref{tab:benchmark_table} shows the percentage of instances where the model-identified main difference matched the human-reported one, with GPT-4o leading at 39\% accuracy. Across all models, performance improved by up to 98\% with accurate edits, a trend also seen in generating all differences captions. `Remove' edits had the lowest performance, with accuracy dropping by up to 50\% compared to the best-performing `Replace' edits.

\sparagraph{0.8ex}{All Differences Captions:} Table \ref{tab:benchmark_table} shows that GPT-4o and Qwen2.5-VL achieves the highest Model Precision (MP) at 12\%, while Qwen2.5-VL demonstrates the lowest Hallucination Rate (HR) at 58\%, along with notable improvements in soft metrics, suggesting confusion between source and target objects. Overall performance remains suboptimal, as model predictions often misalign with human annotations. On average, models describe 1-3.1 differences per image, whereas human annotators identified six differences on average. This gap highlights models’ difficulty in capturing subtle differences and their tendency to overlook details or introduce hallucinated changes. 

Additionally, we observed that models tend to hallucinate less where the edits are accurately performed, leading to a 29\% improvement in HR and a threefold increase in MP across all models.

\textbf{Models vary significantly in predicting no differences between images.} For example, Gemini-1.5 predicts no differences in 24\% of the examples, compared to only 0.3\% for GPT-4o. Gemini-1.5’s higher rate of ``no difference'' predictions lowers its HR but causes it to identify fewer differences than GPT-4o, which detects 80\% more differences while keeping a lower HR. When the edit is contextually consistent, most models predict no differences up to 3 times more often, suggesting they are more sensitive to semantic flaws than pixel-level ones.

\textbf{Models struggle with Remove edits while excelling in Add edits.} All models perform best on Add edits and worst on Remove edits, with Model Precision (MP) differing by up to 2.7x. The Hallucination Rate (HR) for Remove edits is significantly worse, increased up to 50\% compared to Add edits.

\textbf{Models are sensitive to scene complexity (i.e., the number of objects)}. Figure \ref{fig:number_of_objects_comparison} in the Appendix shows that as the number of objects increases, all models exhibit declining precision and rising hallucination rates. GPT-4 and GPT-4-turbo, in particular, struggle more with complex scenes, showing sharp increases in hallucinations. While Gemini-1.5 and GPT-4o also degrade, their decline is less steep.  This trend was not observed in the Edit Inspector questions (Yes/No questions).


\section{New Methods}
\label{sec:new_methods}
To tackle the challenges models face in generating accurate difference captions and detecting unintended artifacts, we developed a zero-shot pipeline for producing detailed, instruction-grounded captions (\Cref{sec:difference_caption_pipeline}) and an artifact detection method using segmentation model probabilities (\Cref{sec:artifacts_detection}). Our methods are competitive with the best models, and in the main difference generation task outperform them by 36\% margin.
\subsection{Difference Caption Pipeline}
\label{sec:difference_caption_pipeline}
\begin{figure*}[!h]
    \centering
    \includegraphics[width=1\textwidth]{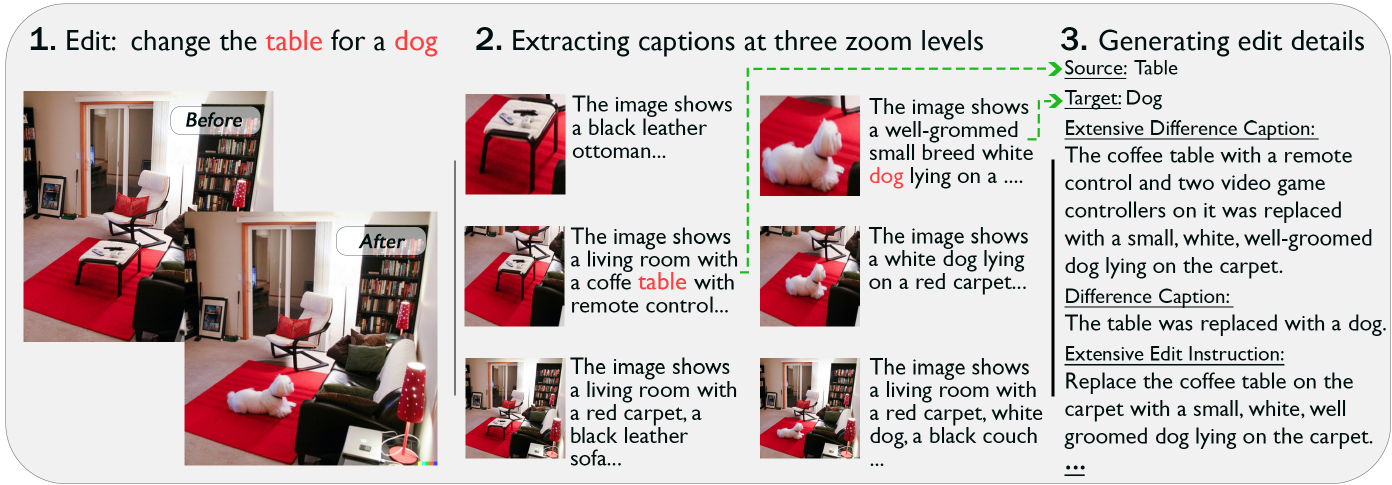}
    \caption{Example of our pipeline generating an instruction-grounded difference caption with rich metadata. Edit images are split into three zoom levels, with Gemini extracting and prioritizing captions to generate the metadata.}
    \label{fig:pipeline}
\end{figure*}
Our pipeline generates detailed, instruction-grounded difference captions and rich metadata by selecting image captions of the edited object area that align with the edit instructions. \textbf{It achieves 75\% accuracy in describing the main edit, surpassing GPT-4o's 39\% accuracy.}

The pipeline extracts image captions at three zoom levels around the edit in-painting mask for both the source and target images. We then select the captions that best match the edit instructions, measured by the number of shared nouns or their synonyms using WordNet \cite{wordnet}. Using these grounded captions and the edit instruction, we employ a one-shot prompt with GPT-4 \cite{openai2024gpt4technicalreport} to generate a detailed difference caption with metadata, as shown in \Cref{fig:pipeline}. 

We found this method most effective for generating a main difference caption. Other methods, such as asking object-specific questions or requesting long image descriptions, often resulted in significant hallucinations and incorrect or biased descriptions. This issue persisted with different visual backbones, such as GPT-4 \cite{openai2024gpt4technicalreport}, LLAVA 1.5 \cite{liu2024improvedbaselinesvisualinstruction}, etc. Integrating human instructions with edited area descriptions allow information sharing as seen in Figures \ref{fig:cake-plate-before}, \ref{fig:cut-pineapple-before}.

\subsection{Artifact Detection}
\label{sec:artifacts_detection}
We developed two artifact detection methods using the extracted metadata from our pipeline. The first method uses the Detic model \cite{zhou2022detecting} to analyze the segmentation probability of each object intersected by the edit mask. A drop of the probability score by more than 4\% as a result of the edit is considered an artifact.

The second method identifies elements that intersect with the mask area, have disappeared from the image, and do not overlap with the edited object's bounding box. This often occurs when the mask is large, but the edited object is small.

\begin{figure*}[h]
    \centering
    \includegraphics[width=1\textwidth]{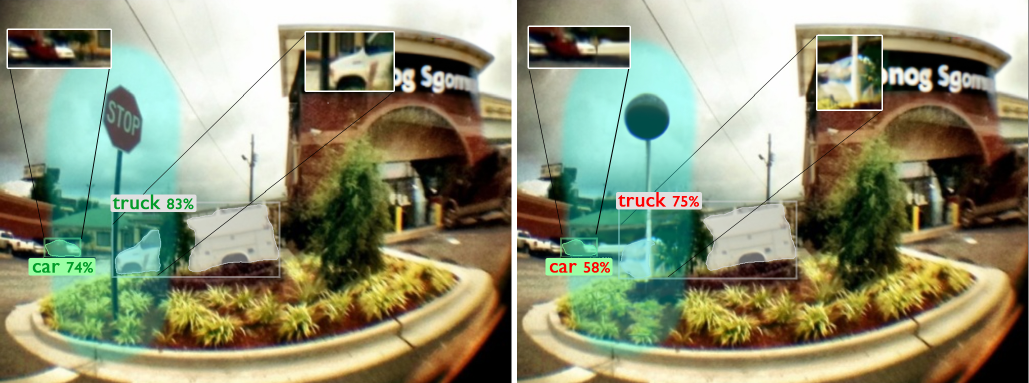}
    \caption{The first method for detecting artifacts using the Detic model for the edit ``turn the stop sign to a lollipop''. Comparing Detic probabilities for objects intersecting the turquoise in-painting mask between the pre-edit (left) and post-edit (right) images reveals two artifacts, the truck and small car, whose probability drops exceeds our threshold.}
    \label{fig:artifacts_detection_example}
\end{figure*}

Combined, \textbf{our methods achieve 64\% balanced accuracy in detecting ``Significant'' artifacts, competitive only with GPT-4o scoring 65.7\%.} \Cref{fig:artifacts_detection_example} shows the first artifact detection method. If the small car intersecting with the in-painting area had been unintentionally removed, it would illustrate the second method.

An oracle that combines the optimal predictions from GPT-4o and our artifact detection method reaches a score of 86.8\% with 100\% precision. \textbf{This indicates that our artifact method and GPT-4o correctly classify different sets of examples.}

\section{Model Supervision}
\label{sec:supervision}
We introduce an end-to-end fine-tuned LLaVA (Language-Vision Alignment) model that rivals much larger models in performance. It offers edit evaluation abilities equivalent to SoTA models while significantly reducing computational costs, providing an efficient solution for AI-generated image edit evaluation.

We trained the model using the MagicBrush train set consists of 8,808 edits. A balanced set of 5,422 edits was used for artifact detection. For edit accuracy and caption generation, the entire set was used, and 31,059 training instances were produced using the two augmentation methods described below. 
Further details are provided in Appendix~\ref{sec:supervision_details}.
\subsection{Negative Edit Augmentation}
\label{sec:negative_edit_augmenation}
This method generates negative edits by selecting a deceptive target object and producing corresponding metadata, including instructions and difference captions. In \Cref{fig:augmentation_example}, a similarly sized scene object (an umbrella) was chosen as the deceptive target, and new metadata was generated using GPT-3.5 with few-shot prompting. For Add and Replace edits, the deceptive object is a visually similar absent object, like a cactus instead of a potted plant. For Change Attribute edits, attributes are modified, like altering a coat’s color from blue to red.
\subsection{Reverse Edit Augmentation}
\label{sec:reverse_edit_augmenation}
This augmentation focuses on reversing the edit using few-shot prompts with GPT-3.5. Add edits are changed to Remove edits, Replace edits involve switching the source and target objects, and Change Attribute edits reverse the attribute modification. For example, in Figure~\ref{fig:augmentation_example}, the edit ``Remove one potted plant'' is reversed to ``Add one potted plant.'' Applied on top of the negative augmentation, this process expands the dataset fourfold, providing comprehensive training data for our model.
\subsection{Supervision Results}
\label{sec:supervision_results}
\textbf{Our model demonstrates competitive performance against SoTA VLMs.} As shown in Table~\ref{tab:benchmark_table}, it outperforms Gemini, GPT-4, and GPT-4-turbo in artifact detection, with only Gemini-1.5 (58.5\%) and GPT-4o (65.7\%) performing better. For Edit Accuracy, it achieves 67.2\%, surpassing Gemini (49.9\%) and GPT-4 Turbo. It also maintains competitive performance in the Difference Caption Accuracy (54.5\%), surpassing Gemini model (53.9\%). These results validate our augmentation methods and highlight the value of our training data.


\section{Additional Editing Methods}
\label{sec:additional_edit_models}
To keep up with the latest image editing models and provide more robust evaluations of models as Edit Inspectors, we used 100 edits from the MagicBrush test set to generate mask-guided image edits using UltraEdit \cite{zhao2024ultraedit} and Imagen3 \cite{imagenteamgoogle2024imagen3}.

We annotated these edits using the same methods described in Section \ref{sec:human_annotation}, with the distribution shown in Table \ref{tab:additional_models_distribution}. Comparing the human labels, we found that MagicBrush edits achieved the highest overall accuracy (85.3\%). Imagen3 had the highest “Visual Consistency” (37.7\%), “Technical Precision” (61.6\%), and the fewest edits without artifacts (10.10\%). UltraEdit showed the lowest “Visual Consistency” (10.67\%) and lowest accuracy rate (31\%), highlighting variation in edit quality across models. These findings show that the latest image editing models exhibit notable weaknesses.

Tables \ref{tab:ultraedit_100_benchmark_table}, \ref{tab:imagen3_100_benchmark_table}, and \ref{tab:magicbrush_100_benchmark_table} show model performance as Edit Inspectors. We observe a consistent decline in performance on the Edit Inspectors Questions for both Imagen3 and UltraEdit edits. GPT-4o remains relatively strong in identifying main differences, but its performance, as well as that of other models, drops across core quality dimensions such as accuracy, technical precision, etc. In contrast, performance on Difference Caption Generation remains comparatively stable. These results further support our observation that current models struggle to evaluate edits comprehensively and frequently hallucinate when describing changes.
\section{Related Work}
\label{sec:related_work}
Recent advances in text-guided image editing enable modifications via natural language \cite{sheynin2023emu, he2024llms, wu2023visual, cui2023chatedit}, with some models supporting multi-turn refinement \cite{cui2023chatedit}. Others use spatial masks for precise, localized edits \cite{Avrahami_2022, nichol2022glide, Editbench}, which offer better control than text-only methods \cite{Editbench, Magicbrush}.

Edit quality is often measured using pixel-level similarity (L1/L2 norms) and CLIP-based cosine similarity \cite{CLIP2021}. However, these metrics poorly align with human judgment \cite{editval}, offering only quantitative scores without qualitative insights.

Image editing benchmarks like EditBench \cite{Editbench} and EditVal \cite{editval} assess editing models through automatic and human evaluations, focusing on instruction adherence and object or scene preservation. In contrast, our work evaluates models as edit inspectors on overlooked edit aspects such as scene integration, pixel-level issues, and artifact detection. We also introduce the category ``Accurate, But Unexpected'' to capture technically correct edits that deviate from user expectations and collect textual feedback and detailed difference captions to provide deeper insights into edit quality.


\section{Conclusion}
\label{sec:conclusion}
In this work, we introduce \shortpapername, a public benchmark for assessing
evaluators of text-guided image edits across several dimensions: accuracy,
artifact detection, visual quality, seamless integration with the image scene, adherence to common sense, and the ability to describe edit-induced changes. Using the EditInspector benchmark, we show that state-of-the-art vision and language models perform poorly as edit inspectors and cannot effectively assess edits. To address these limitations, we propose two novel methods and fine-tune an edit inspector that outperforms these models. We hope that our benchmark and proposed methods will drive advancements in edit evaluation and inspire further research in this domain.

\section{Future Work}
\label{sec:future_work}
Future work can refine difference caption generation and explore new approaches to address existing model limitations. 
Additionally, several directions could further expand the benchmark’s coverage and evaluation capabilities:
\begin{itemize}
    \item \textbf{Incorporating complex multi-object and multi-operation edits:} Expanding the benchmark to edits involving multiple objects and operations, such as simultaneously adding new objects while removing others, would enable a broader assessment of model generalization.
    \item \textbf{Supporting multi-turn edit evaluation:} Extending the benchmark to include sequential edits would allow evaluation of models' ability to maintain visual and semantic consistency across multiple editing steps.
\end{itemize}

\section{Limitations}

Our benchmark is based exclusively on the MagicBrush dataset for evaluating edits, which, while covering diverse scenarios, is limited to natural images and mask-guided edits. Recent studies have shown promising results with free-text methods \cite{sheynin2023emu} and growing interest in editing of synthetic images. Additionally, the distribution of edit types in the test set reflects the natural distribution of human edits from the MagicBrush dataset, as determined by a human study. While this mirrors real-world editing trends, it may not equally represent all edit types. These limitations highlight distinct research directions that could be explored independently of our current work.

\bibliography{custom}

\appendix
\section{Appendix}
\subsection{Common Caption Comparison Metrics}
\label{sec:common_metrics_for_caption_comperasion_overview}
Common metrics for comparing image captions, such as BLEU, METEOR, and ROUGE, rely on N-gram overlaps between generated and reference texts. However, they fall short of our core requirement to ensure accurate alignment between the edited objects and actions described in the captions. As shown in Table~\ref{tab:model_and_lanuge_metrics_performance}, while these metrics suggest that GPT-4 generates captions most similar to the ground truth, in practice, it is the least accurate model, exhibiting the highest hallucination rate and the largest number of average changes detected. Below, we provide a brief explanation of these metrics, followed by several scenarios illustrating their limitations in effectively evaluating difference captions.
\begin{table*}[tp]
    \centering
    \small
    \begin{tabular}{lcccc}
        \toprule
                                 & Gemini-1.5       & GPT-4         & GPT-4o          & GPT-4 Turbo  \\
        \midrule
        \multicolumn{5}{c}{\textbf{Differences Caption Generation}} \\
        \midrule
        Main Difference             & 31\%           & 27\%        & \textbf{39\%} & 24\%           \\
        MP                          & 8\%              & 8\%           & \textbf{12\%}   & 8\%            \\
        HR                          & 67\%             & 78\%          & \textbf{60\%}   & 75\%           \\
        \midrule
        METEOR                      & 0.11            & \textbf{0.22}          & 0.19            & 0.19           \\
        ROUGE-1                     & 0.15            & \textbf{0.36}          & 0.29            & 0.30           \\
        ROUGE-2                     & 0.04            & \textbf{0.09}          & 0.08            & 0.07           \\
        BLEU                        & 0.01            & 0.02         &      \textbf{0.03}        & 0.02           \\
        \bottomrule
    \end{tabular}
   \caption{Comparison of models on the Difference Caption Generation task. GPT-4 achieves the best results on METEOR, ROUGE-1, and ROUGE-2 metrics, while GPT-4o ranks highest in BLEU.}
    \label{tab:model_and_lanuge_metrics_performance}
\end{table*}

\begin{itemize}
    \item BLEU: Computes the number of matches in unigrams, bigrams, trigrams, and 4-grams between generated and reference text. Includes a brevity penalty to discourage shorter outputs.
    \item ROUGE: ROUGE-1 calculates the F1 score for unigrams. ROUGE-2 calculates the F1 score for bigrams.
    \item METEOR: Incorporates features such as stemming, synonym matching, and paraphrase recognition. Computes the unigram F1 score.
    \item CIDEr: Measures the similarity between generated and reference captions using TF-IDF weighted n-grams (unigrams to 4-grams). Emphasizes consensus between generated captions and multiple human references while penalizing overuse of common n-grams.
\end{itemize}
Although these metrics are widely used in image captioning, they have severe limitations when evaluating difference captions for image edits.

\xhdr{Miss Weighting the Edited Objects and Actions}
These metrics struggle to differentiate between critical objects and less significant words in the context of difference captions. For instance, consider the ground truth caption: "The main difference between the two images is the first image has a blue vase and the second image a brown vase." If the generated caption states, "The main difference between the two images is the first image has a squirrel and the second image does not," linguistic metrics might still assign relatively high scores (e.g., BLEU: 0.68, ROUGE-1 Recall: 0.81, METEOR: 0.78) due to superficial word overlaps. However, these scores fail to reflect the semantic misalignment between the captions. In contrast, our proposed metric assigns a score of 0, accurately reflecting the discrepancy in the identified edited object and action.

\xhdr{Accounting for Unchanged Objects, Varying Length, and Stylistic Differences}
Conventional metrics often penalize captions that include mentions of unchanged objects, vary in length, or differ stylistically, even when accurately describing the detected changes. For instance, consider the generated caption: "The difference between the two images is that the first image has a blue vase. The second image has a blue vase and a squirrel next to it." Our metric would assign this caption a perfect score of 1, as it correctly identifies the key difference (the addition of the squirrel) in alignment with the ground truth caption: "A brown squirrel was added to the image." In contrast, linguistic metrics would score close to 0 due to the inclusion of details about the unchanged "blue vase" and penalties for variations in length and phrasing. This demonstrates the robustness of our metric in handling linguistic variability while focusing on the accuracy of detected changes.

\xhdr{Capturing the Order of Edits}
The above mentioned metrics overlook the importance of edit sequence order. For instance, consider the ground truth captions: "In the first image, the tree was removed, and a new flowerbed was added" and the generated caption "In the first image, the flowerbed was removed, and a new tree was added." Although both captions involve the same objects (tree and flowerbed) and actions (added and removed), the sequence of edits conveys entirely different meanings. The n-gram based metrics would assign high scores to these captions because they mention the same words (objects and actions), regardless of their order, failing to penalize semantic misalignment. In contrast, our metric explicitly evaluates the edit sequence order, ensuring that generated captions accurately reflect the correct sequence of changes.

\subsection{Mathematical Explanation of Metrics}
\label{sec:metics_mathematical_explenations}
We evaluate model performance on \textbf{all differences captions} using two metrics: \textbf{Model Precision (MP)} and \textbf{Hallucination Rate (HR)}. These are computed based on Difference Triplets (DTs), defined as:
\[\text{DT} = (\text{source object}, \text{target object}, \text{action type}),\]
where \textit{source object} is the original object affected by the edit, \textit{target object} is the resulting object of the edit, and \textit{action type} is the type of edit (e.g., "add," "remove," "replace").
\textbf{Model Precision (MP):} Measures the percentage of human-annotated DTs (\( \mathcal{H} \)) matched by model-detected DTs (\( \mathcal{M} \)):
\[\text{MP} = \frac{\lvert \mathcal{H} \cap \mathcal{M} \rvert}{\lvert \mathcal{H} \rvert} \times 100,\]
where \( \lvert \mathcal{H} \cap \mathcal{M} \rvert \) is the number of matched DTs, and \( \lvert \mathcal{H} \rvert \) is the total human-annotated DTs.
\textbf{Hallucination Rate (HR):} Measures the percentage of model-detected DTs (\( \mathcal{M} \)) not matching any human-annotated DTs (\( \mathcal{H} \)):
\[\text{HR} = \frac{\lvert \mathcal{M} \setminus \mathcal{H} \rvert}{\lvert \mathcal{M} \rvert} \times 100,\]
where \( \lvert \mathcal{M} \setminus \mathcal{H} \rvert \) is the number of hallucinated DTs, and \( \lvert \mathcal{M} \rvert \) is the total model-detected DTs.
\textbf{Soft Metrics:} MP\textsubscript{soft} and HR\textsubscript{soft} allow matches when source and target objects in DTs are reversed:
\[\text{MP\textsubscript{soft}} = \frac{\lvert \mathcal{H}_{\text{soft}} \cap \mathcal{M} \rvert}{\lvert \mathcal{H} \rvert} \times 100, \quad \]
\[\text{HR\textsubscript{soft}} = \frac{\lvert \mathcal{M} \setminus \mathcal{H}_{\text{soft}} \rvert}{\lvert \mathcal{M} \rvert} \times 100.\]
\textbf{Matching Criteria:} A DT match requires identical \textit{action type} and similar \textit{source}/\textit{target objects} (assessed by GPT-4). Relaxed matching (\( \mathcal{H}_{\text{soft}} \)) accounts for reversed source and target objects.
\subsection{Additional Editing Methods}
\label{sec:additional_editing_models}
\begin{table*}[tp]
    \centering
    \small
    \begin{tabular}{lccc}
        \toprule
        Category                          & Imagen3 (\%) & UltraEdit (\%) & MagicBrush (\%) \\
        \midrule
        \makecell{Accuracy \\ Level}       
            & \makecell[l]{Accurate: 15.82 \\ Accurate Unexpected: 46.46 \\ Inaccurate: 18.86 \\ Inaccurate  Reflects: 18.86}
            & \makecell[l]{Accurate: 6.3 \\ Accurate Unexpected: 46 \\ Inaccurate: 31 \\ Inaccurate Reflects: 16.7} 
            & \makecell[l]{Accurate: 16 \\ Accurate Unexpected: 69.3 \\ Inaccurate: 7 \\ Inaccurate Reflects: 7.7} \\
        \midrule
        \makecell{Artifacts \\ Level}      
            & \makecell[l]{Significant: 36.7 \\ Mild: 53.2 \\ No Artifact: 10.1} 
            & \makecell[l]{Significant: 29.3 \\ Mild: 65 \\ No Artifact: 5.7} 
            & \makecell[l]{Significant: 36.7 \\ Mild: 58 \\ No Artifact: 5.3} \\
        \midrule
        \makecell{Technical \\ Precision}  
            & \makecell[l]{Yes: 61.6 \\ No: 38.3} 
            & \makecell[l]{Yes: 37.6 \\ No: 62.3} 
            & \makecell[l]{Yes: 39 \\ No: 61} \\
        \midrule
        \makecell{Visual \\ Consistency}   
            & \makecell[l]{Yes: 37.7 \\ No: 62.3} 
            & \makecell[l]{Yes: 10.7 \\ No: 89.3} 
            & \makecell[l]{Yes: 22 \\ No: 78} \\
        \bottomrule
    \end{tabular}
    \caption{Distribution of annotation values across categories for Imagen3, UltraEdit, and MagicBrush. The table summarizes the percentage breakdown for each evaluation category. Notably, MagicBrush had the highest percentage of overall Accurate edits, while Imagen3 showed the strongest performance in Technical Precision.}
    \label{tab:additional_models_distribution}
\end{table*}

\xhdr{UltraEdit Annotation Agreement}
\textbf{Accuracy} stood out with a high complete majority rate (78\%) and the highest average agreement rate (92.67\%). For the more detailed \textbf{accuracy levels}, majority vote was reached in 81\% of cases, though the complete majority rate dropped to 53\%, with an average agreement rate of 95\% among those with a majority vote.

In the \textbf{artifacts} category, annotators reached a 64\% complete majority rate, with an average agreement rate of 88\%. The more granular \textbf{artifact levels} followed a similar pattern, with 98\% majority vote, 52\% complete majority rate, and 82.67\% average agreement.

\textbf{Technical Precision} showed a complete majority in only 36\% of cases, with a moderate average agreement rate of 78.67\%. Finally, \textbf{Visual Consistency} achieved a strong 91\% complete majority rate, and a 73\% average agreement rate, reflecting consistent annotator judgment in this category.

\xhdr{Imagen3 Annotation Agreement}
\textbf{Accuracy} achieved a complete majority rate of 83.84\% and the highest average agreement rate (94.61\%) across all categories. For the more detailed \textbf{accuracy levels}, majority vote was reached in 95.96\% of cases, though the complete majority rate dropped to 36.36\%, with an average agreement rate of 76.09\%.

In the \textbf{artifacts} category, annotators reached a 58.59\% majority vote and a 64\% complete majority rate, with an average agreement rate of 86.20\%. The more granular \textbf{artifact levels} showed a 94.95\% majority vote, 46.46\% complete majority rate, and 78.79\% average agreement.

\textbf{Technical Precision} had a complete majority rate of 44. 44\% and an average agreement rate of 81.48\%. \textbf{Visual Consistency} achieved a 45.45\% complete majority rate and 81.82\% average agreement, indicating relatively stable annotator consensus in this category.

\xhdr{MagicBrush Annotation Agreement}
\textbf{Accuracy} levels in MagicBrush exhibited strong annotator alignment, with a 95\% majority vote rate and an average agreement rate of 80.67\%, though the complete majority rate was moderate at 52\%.

For the \textbf{artifacts} levels, annotators reached a 99\% majority vote, with an average agreement rate of 83\% and a complete majority rate of 51\%, closely mirroring previous patterns observed in Imagen3.

\textbf{Technical Precision} (annotated as good quality) showed a complete majority in 41\% of cases, with an average agreement rate of 80.33\%.

Finally, \textbf{Visual Consistency} achieved a 57\% complete majority rate and the highest average agreement rate in MagicBrush at 85.67\%, suggesting relatively consistent annotator judgments in this category.

\begin{table*}[tp]
    \centering
    \small
    \begin{tabular}{lcccccccc}
        \toprule
                                 & GPT-4         & GPT-4o          & \makecell{GPT-4 \\ Turbo}  & \makecell{Qwen2.5 \\ VL}  & \makecell{InternVL3} & LLaVA  &  \makecell{LLaVA \\ (Supervised)}\\
        \midrule
        \multicolumn{6}{c}{\textbf{Edit Inspectors Questions}} \\
        \midrule
        \makecell[l]{Accuracy}                    & \textbf{63\%}       & 51.8\%        & 57.5\%  & 62.1\%          & 54.4\%     & 52.6\%  & 54.3\% \\
        \makecell[l]{Contextual  \\  Consistency} & 48.5\%        & 41.6\%      & 37.1\%   & 58.2\%       & \textbf{61.9\%}     & -  & - \\
        \makecell[l]{Technical  \\  Precision}    & \textbf{53.4\%}        & 46.4\%           & 48.5\%    & 43.8\%         & 46.2\%     & -  &  - \\
        \makecell[l]{Artifacts}                   & 42.9\%     & \textbf{56.3\%}        & 51\%   & 50\%       & 41\%  &    52.7\%         & 55.6\%\\
        \makecell[l]{Difference  \\ Caption Acc}  & 55.6\%        & 55.6\%          & \textbf{57.5\%}     & 54.4\%          & 44.4\%   & 50\%  & 48.1\%\\
        \midrule
        \multicolumn{6}{c}{\textbf{Differences Caption Generation}} \\
        \midrule
        \makecell[l]{Main  \\  Difference}            & 36\%          & \textbf{44\%}   & 36\%      & 8\%          & 22\%    & 4\%       & 9\% \\
                            MP                        & 7\%           & \textbf{9\%}     & 7\%       & 7\%          & 8\%     & -      & - \\
                            MP\textsubscript{soft}    & 9\%           & 11\%     & 8\%       & 9\%         & \textbf{12\%}     & -     & - \\
                            HR                        & 82\%          & \textbf{74\%}    & 80\%      & 84\%         & 90\%     & -     & - \\
                            HR\textsubscript{soft}    & 80\%          & \textbf{68\%}    & 77\%     & 80\%         & 86\%    & -      & - \\
                            \midrule
                            Avg. Diff                 & 2.5           & 1.9         & 1.5    & 2.5          & \textbf{3.4}      & -      & - \\
                            No Diffs                  & 0.8\%           & 0\%           & \textbf{4.5\%}    & 0.8\%          & 3.2\%  & -      & - \\
        \bottomrule
    \end{tabular}
    \caption{Models performance on UltraEdit edits across models Edit Inspectors questions and Difference Caption Generation. The first section reports binary question accuracy on core evaluation criteria (Accuracy, Contextual Consistency, Technical Precision, and Artifacts). The second section presents difference caption metrics: percentage of predicted main difference captions correctly describing the main difference, hallucination rates (HR), and average number of predicted differences. \textit{Avg. Diff} indicates the mean number of differences per edit, and \textit{No Diffs} reports the percentage of edits with no predicted differences.}
    \label{tab:ultraedit_100_benchmark_table}
\end{table*}

\begin{table*}[tp]
    \centering
    \small
    \begin{tabular}{lcccccccccc}
        \toprule
                                     & GPT-4         & GPT-4o          & \makecell{GPT-4 \\ Turbo}  & \makecell{Qwen2.5 \\ VL}  & \makecell{InternVL3} & LLaVA  &  \makecell{LLaVA \\ (Supervised)}\\
        \midrule
        \multicolumn{6}{c}{\textbf{Edit Inspectors Questions}} \\
        \midrule
        \makecell[l]{Accuracy}                    & \textbf{55.3\%}        & 49.9\%          & 49.2\%   & 52.9\%          & 51.5\%      & 54.4\%    & 58.8\% \\
        \makecell[l]{Contextual  \\  Consistency} & 47.7\%        & 49.5\%          & \textbf{55.6\%}   & 42.1\%          & 48.9\%       & -    & - \\
        \makecell[l]{Technical  \\  Precision}    & 51.4\%        & 50.9\%          & 49.1\%   & 47.5\%          & \textbf{54.5\%}     & -    & - \\
        \makecell[l]{Artifacts}                   & 47.1\%        & \textbf{56.5\%}          & 49.5\%   & 50.0\%          & 48.6\%     & 43.6\%    & 48.8\% \\
        \makecell[l]{Difference  \\ Caption Acc}  & 54.2\%        & \textbf{58.2\%}          & 55.9\%   & 51.8\%          & 50.8\%     & 50\%  & 53.9\% \\
        \midrule
        \multicolumn{6}{c}{\textbf{Differences Caption Generation}} \\
        \midrule
        \makecell[l]{Main  \\  Difference}        & 29\%   & \textbf{44\%}   & 27\%      & 29\%    & 13\%  & 4\%     & 11\% \\
        MP                          & 9\%           & \textbf{11\%}      & 8\%     & 10\%          & 9\%    &  -      & -\\
        MP\textsubscript{soft}      & 11\%          & \textbf{12\%}      & 9\%     & \textbf{12\%}          & \textbf{12\%}    &  -     &  - \\
        HR                          & 78\%          & \textbf{73\%}      & 82\%    & \textbf{73\%}          & 92\%   &  -      &  - \\
        HR\textsubscript{soft}      & 73\%          & 70\%      & 80\%    & \textbf{67\%}          & 88\%     &  -      &  - \\
        \midrule
        Avg. Diff                   & 2.5         & 1.9           & 1.5    & 1.5          & \textbf{3.2}     & -      & - \\
        No Diffs                    & 0.8\%         & 0\%           & \textbf{4.5\%}    & 1.2\%          & 4\%      & -      & - \\
        \bottomrule
    \end{tabular}
    \caption{Models performance on Imagen3 edits Edit Inspectors questions and Difference Caption Generation. The first section reports binary question accuracy on core evaluation criteria (Accuracy, Contextual Consistency, Technical Precision, and Artifacts). The second section presents difference caption metrics: percentage of predicted main difference captions correctly describing the main difference, hallucination rates (HR), and average number of predicted differences. \textit{Avg. Diff} indicates the mean number of differences per edit, and \textit{No Diffs} reports the percentage of edits with no predicted differences.}
    \label{tab:imagen3_100_benchmark_table}
\end{table*}


\begin{table*}[tp]
    \centering
    \small
    \begin{tabular}{lcccccccc}
        \toprule
                                     & GPT-4         & GPT-4o          & \makecell{GPT-4 \\ Turbo}  & \makecell{Qwen2.5 \\ VL}  & \makecell{InternVL3} & LLaVA  &  \makecell{LLaVA \\ (Supervised)}\\
        \midrule
        \multicolumn{6}{c}{\textbf{Edit Inspectors Questions}} \\
        \midrule
        \makecell[l]{Accuracy}                    & \textbf{73.7\%}        & 71\%            & 60\%     & 64\%          & 63.6\%    & 58.5\% & 62.3\%  \\
        \makecell[l]{Contextual  \\  Consistency} & 48.2\%        & 56.7\%          & \textbf{59.7\%}   & 38.4\%          & 45.6\%     & -    & - \\
        \makecell[l]{Technical  \\  Precision}    & 49.2\%        & 47.6\%       &   \textbf{49.8\%}   & 47.5\%          & 42.6\%    & -    & - \\
        \makecell[l]{Artifacts}                   & 51.6\%        & \textbf{60.6\%}          & 54.1\%   & 50\%          & 52.5\%     & 44.8\%    & 56.2\% \\
        \makecell[l]{Difference  \\ Caption Acc}  &\textbf{63.2\%}        & 60.7\%          & 61.1\%   & 59.8\%          & 58.4\%    & 50\%  & 46.5\% \\
        \midrule
        \multicolumn{6}{c}{\textbf{Differences Caption Generation}} \\
        \midrule
        \makecell[l]{Main  \\  Difference}        & 34\%   & \textbf{50\%}   & 34\%      & 37\%    & 25\%    & 5\%     & 12\% \\
        MP                          & 8\%           & \textbf{11\%}      & 7\%     & \textbf{11\%}  & 8\%     & -      & - \\
        MP\textsubscript{soft}      & 8\%          & \textbf{12\%}       & 8\%     & 1\%           & 11\%     & -      & - \\
        HR                          & 75\%          & \textbf{61\%}      & 78\%    & 63\%          & 88\%     & -      & - \\
        HR\textsubscript{soft}      & 74\%          & \textbf{57\%}      & 75\%    & 59\%          & 83\%     & -      & - \\
        \midrule
        Avg. Diff                   & 2.5         & 1.9           & 1.5    & 1.5          & \textbf{3.4}       & -      & - \\
        No Diffs                    & 0.8\%         & 0\%           & \textbf{4.5\%}    & 0.6\%          & 2.3\%     & -      & - \\
        \bottomrule
    \end{tabular}
    \caption{Models performance on MagicBrush edits Edit Inspectors questions and Difference Caption Generation. The first section reports binary question accuracy on core evaluation criteria (Accuracy, Contextual Consistency, Technical Precision, and Artifacts). The second section presents difference caption metrics: percentage of predicted main difference captions correctly describing the main difference, hallucination rates (HR), and average number of predicted differences. \textit{Avg. Diff} indicates the mean number of differences per edit, and \textit{No Diffs} reports the percentage of edits with no predicted differences.}
    \label{tab:magicbrush_100_benchmark_table}
\end{table*}

\subsection{Metrics Example}
\label{sec:metics_example}
We calculate the MP and HR metrics using Figure~\ref{fig:models_results_example} GPT-4o and the human-annotated difference caption. The ground truth lists the following human-annotated differences $(\mathcal{H})$:
\[
\begin{aligned}
    &(\text{carpet floor}, \text{wooden floor}, \text{Replace}), \\
    &(\text{None}, \text{door}, \text{Add}), \\
    &(\text{fridge bottom}, \text{extended fridge bottom}, \text{Change}), \\
    &(\text{yellow box}, \text{extended yellow box}, \text{Change}), \\
    &(\text{yellow box text}, \text{None}, \text{Remove}), \\
    &(\text{text}, \text{image}, \text{Replace})
\end{aligned}
\]
GPT-4o detects only one difference:
\[\mathcal{M} = \{(\text{carpet floor}, \text{wooden floor}, \text{Replace})\}.\]
\textbf{Model Precision (MP):}  
Model Precision (MP) measures the percentage of human-annotated DTs (\( \mathcal{H} \)) matched by model-detected DTs (\( \mathcal{M} \)):
\[\text{MP} = \frac{\lvert \mathcal{H} \cap \mathcal{M} \rvert}{\lvert \mathcal{H} \rvert} \times 100.\]
The only match between \( \mathcal{H} \) and \( \mathcal{M} \) is:
\[(\text{carpet floor}, \text{wooden floor}, \text{Replace})\]
Therefore:
\[\lvert \mathcal{H} \cap \mathcal{M} \rvert = 1, \quad \lvert \mathcal{H} \rvert = 6,\]
\[\text{MP} = \frac{1}{6} \times 100 \approx 16.67\%.\]
\textbf{Hallucination Rate (HR):}  
Hallucination Rate (HR) measures the percentage of model-detected DTs (\( \mathcal{M} \)) that do not match any human-annotated DTs (\( \mathcal{H} \)):
\[\text{HR} = \frac{\lvert \mathcal{M} \setminus \mathcal{H} \rvert}{\lvert \mathcal{M} \rvert} \times 100.\]
Here, all model-detected DTs match human-annotated DTs, so:
\[\mathcal{M} \setminus \mathcal{H} = \emptyset, \quad \lvert \mathcal{M} \rvert = 1,\]
\[\text{HR} = \frac{0}{1} \times 100 = 0\%.\]
\[\text{MP} = 16.67\%, \quad \text{HR} = 0\%.\]
\subsection{Detailed Description of New Methods}
\label{sec:new_methods_appendix}
\sparagraph{0.6ex}{Caption Pipeline}
We introduce a structured pipeline to generate difference captions that describe the specific visual changes between the source and edited images. This pipeline combines both image-level and region-level vision-language descriptions with a prompt-based large language model (LLM) querying strategy to produce rich and accurate textual metadata for each edit.

The process begins by extracting bounding boxes from the user-provided edit mask. If a single bounding box is found, we crop the source and target images in two ways: a tight crop and a padded version, where the box is either doubled in size or expanded to cover at least 15\% of the image dimensions. If multiple bounding boxes are present or if cropping fails, we fall back to using the full images.

We then apply a vision-language model (Gemini) to generate descriptions of the cropped and full image regions. For each image, we obtain textual descriptions of the masked region, the full image, and the padded area (if applicable). These serve as candidate visual contexts for the next stage of the pipeline.

To ensure that the most relevant image region is used for caption generation, we apply a noun-based grounding mechanism to select the best image description for each image. We extract nouns from the instruction and compare them—using both exact matches and synonym overlap—to nouns in each candidate region description. If the default crop lacks sufficient alignment, we select the region (tight crop, padded crop, or full image) with the highest degree of noun-level overlap. When no region aligns directly, we fall back to the one that shares object mentions with at least one other candidate. This step ensures that the final prompt is grounded in a region of the image that is semantically aligned with the user instruction.

A predefined prompt template is then populated with the instruction and the selected image descriptions. This prompt is passed to GPT-4, which returns a structured response containing multiple fields: the predicted action type, short and extensive difference captions, a revised instruction, source and target object names, and a brief explanation of the edit.

To ensure reliability, the pipeline includes fallback mechanisms if the model returns an invalid action type (e.g., "None"), reattempting generation with alternative image regions. Only examples with valid, well-formed responses are retained.

\sparagraph{0.6ex}{Artifact Detection - Edit Mask Intersection}
Our segmentation-based artifact detection method analyzes object presence and score variation around the edited region to identify potential unintended artifacts. For each object detected by the Detic model in the source and target images, we compute the intersection between its segmentation mask and the user-provided edit mask. We use the object’s binary mask contours rather than bounding boxes, allowing for precise spatial comparisons. To improve robustness, we perform these comparisons at two different resolutions: the source image size and the edited image size, since mask-based scores can fluctuate slightly (by a few percentage points) when resized.

To reduce noise, we exclude very small regions (objects with less than 2.4\% intersection with the mask) as well as objects that are almost entirely contained within the mask (above 97\% intersection), as these are typically intentional edit targets or too minor to evaluate reliably. In addition, we focus specifically on objects that are partially affected by the edit those whose segmentation masks intersect with the edit region by more than 0\% but less than 40\%. This range helps isolate objects that may have been unintentionally damaged during the edit process, as opposed to ones that were fully changed.

For each object class, we retain all detected instances that meet these intersection criteria. To avoid expensive pixel-level comparisons between non-overlapping objects, we first check whether their bounding boxes intersect. This provides a fast pre-filter, since if the bounding boxes do not overlap, their masks cannot intersect either. Only objects with overlapping bounding boxes are further compared across the source and edited images to measure changes in detection confidence. An object is flagged as containing an artifact if its confidence score drops by more than 4\% after the edit, provided it is not fully masked or too small.

\sparagraph{0.6ex}{Artifact Detection - Unintended Object Additions or Removals}

In addition to evaluating score-based degradations near the edit mask, we introduce a second method designed to detect unintended additions or removals of secondary objects within the mask area. This method specifically applies to Add and Remove edits, where the risk of unintentionally modifying unrelated parts of the scene is higher. Unlike the first method, which looks at changes in detection confidence, this approach focuses on complete object disappearance or appearance that may not be aligned with the instruction.

The process begins by identifying the main edit object bounding boxes from the source and target images using object detection. These bounding boxes are used to define the intended region of change.

Next, we detect all other objects present in the source and edited images, and classify them as secondary objects. We then filter out any objects whose class labels are semantically similar to the main object. This includes direct matches, shared noun forms, and synonym relationships. This filtering step ensures that we focus only on truly unrelated objects that are not supposed to be changed.

From this filtered set of secondary objects, we further isolate those that intersect with the mask area but do not spatially overlap with the main object’s bounding boxes. This spatial condition helps distinguish unintended changes from those that are part of the intended edit.

Finally, we compare the presence of these secondary objects across the source and edited images. For Remove actions, if a secondary object is present in the source image but missing in the edited image, we flag it as an unintended removal. For Add actions, if a new secondary object appears in the edited image that was not in the original, we flag it as an unintended addition. If at least one such change is detected, we mark the edit as containing a secondary artifact.

This method captures a different failure mode than the first: it identifies whole-object additions or losses in the masked area that are not part of the intended instruction. Full implementation details, including semantic class filtering, mask-intersection checks, and bounding box exclusion, are available in our released code.
\subsection{Additional Annotation Information}
\label{sec:additional_annotation_information}
Each image edit was annotated by three annotators, with annotations conducted in batches of 27-54 edits. Annotators were paid at a rate of \$$0.70$ per sample, resulting in an average hourly wage of \$$18$. 

To ensure the quality of annotations, we implemented a qualification test to select quality annotators. We provided detailed instructions, including decision trees that visually guide the answering process. These decision trees were accessible via the user interface (``tree icon''), allowing annotators to follow the guidelines while annotating image edits. 

Additionally, a settings window was available, enabling annotators to customize the UI, including font size, width, and padding, to suit their personal preferences (See Appendix \ref{sec:appendx_annotation_ui}).

To assess annotation consistency, we computed Fleiss’ Kappa for each question. The accuracy question showed moderate agreement ($\kappa > 0.41$), suggesting annotators generally aligned on whether an edit was accurate. The artifact question, artifact severity level, accuracy level, and difference caption accuracy all exhibited fair agreement ($\kappa$ between 0.21 and 0.40), indicating moderate consistency across annotators. In contrast, the visual consistency and technical precision questions exhibited only slight agreement ($\kappa$ between 0.01 and 0.20), highlighting greater subjectivity or ambiguity in how these aspects were interpreted.
\subsection{Tasks Prompts}
\label{sec:appendx_taks_prompts}
Model performance varied greatly with different prompts, suggesting that models may struggle to fully understand the task. We selected prompts that conveyed the user instructions and improved the overall performance.
\begin{itemize}
    \item \textbf{Difference Caption Accuracy Task (Yes/No)}

    You are provided with before and after images of an image edit for the edit instruction "\{\}". Does the difference caption "\{\}" describe the difference between the two images (Answer only Yes/No)?

    \item \textbf{Visual Consistency Task (Yes/No)}
    You are provided with before and after images of an image edit for the edit instruction "\{\}". Is the edited object or its area (in remove/replace actions) consistent with the edit instruction and the image scene in terms of shape, size, brightness, shadows, texture, color, etc. (Answer only Yes/No)?
    \item \textbf{Is Accurate Task}

    You are provided with before and after images of an image edit for the edit instruction "\{\}". Was the edit instruction "\{\}" accurately executed and does it reflect the intended change (Answer only Yes/No)?

    \item \textbf{Artifacts Task}

    You are provided with before and after images of an image edit for the edit instruction "\{\}". Are there any artifacts or alterations in the image not intended to be affected by the edit "\{\}" (Answer only Yes/No)?

    \item \textbf{Technical Precision Task (Yes/No)}

    You are provided with before and after images of an image edit for the edit instruction "\{\}". Does the edited object or its area (in remove/replace actions) maintain the image resolution, exhibit blur, show any smoothness, etc. (Answer only Yes/No)?

    \item \textbf{Generate all differences caption}
    You are provided with before and after images of an image edit. Please describe all the differences between these two images. Focus only on the differences; do not include any irrelevant information. Ignore any style differences between the images, such as changes in artistic style, color grading, or filters.

    \item \textbf{Generate main differences caption}
    Please describe the main difference between the two images.
\end{itemize}

\subsection{Textual Feedback}
\label{sec:textual_feedback}
We compared the predicted feedback from the models with human annotations by using a zero-shot prompt with GPT-4o that determines whether two pieces of feedback share any common points (yielding a simple Yes or No). The models' feedback matched human feedback only in a very small percentage of cases. The contextual consistency feedback shared common points with human feedback in 7\%-28\% of cases, while technical precision feedback did so in 4\%-51\% of instances. 

\subsection{Categories of Feedback Issues}
\label{sec:textual_feedback_categories}

\begin{itemize}
    \item \textbf{Shape/Proportion}: Captures distortions in the shape, size, or proportions of objects.  
    \begin{itemize}
        \item \textit{Keywords}: shape, proportion, size, distorted, too big, too small
        \item \textit{Example}: "The bird has an odd shape and is also yellow."
    \end{itemize}

    \item \textbf{Blur/Fuzziness}: Deals with visual issues related to blurred or unclear edges, lack of sharpness, and fuzziness.  
    \begin{itemize}
        \item \textit{Keywords}: blurry, fuzzy, smudged, blurred edges, not clear
        \item \textit{Example}: "The cat’s fur is smoothened and texture is changed."
    \end{itemize}

    \item \textbf{Texture}: Focuses on objects with unrealistic or unnatural textures, often described as too smooth or grainy.  
    \begin{itemize}
        \item \textit{Keywords}: texture, smooth, grainy, patchy, unnatural
        \item \textit{Example}: "The building texture is unnatural."
    \end{itemize}

    \item \textbf{Lighting/Brightness}: Involves issues where shadows are inconsistent or missing, or where lighting is overexposed or underexposed.  
    \begin{itemize}
        \item \textit{Keywords}: shadows, lighting, brightness, overexposed, underexposed
        \item \textit{Example}: "The white bright part on the pan gives it an unrealistic look."
    \end{itemize}

    \item \textbf{Color}: Captures cases where colors are over-saturated, under-saturated, or do not align with the scene.  
    \begin{itemize}
        \item \textit{Keywords}: color, too bright, saturated, unnatural color
        \item \textit{Example}: "The fox is bright and inconsistent with the rest of the image."
    \end{itemize}

    \item \textbf{Unreal/Artificial Look}: Describes objects that appear cartoonish, toy-like, or overly artificial, failing to blend with the rest of the scene.  
    \begin{itemize}
        \item \textit{Keywords}: cartoon, toy, artificial, fake, graphical
        \item \textit{Example}: "The helicopter's texture resembles a toy."
    \end{itemize}

    \item \textbf{Placement}: Refers to objects that are misaligned or incorrectly oriented in the scene.  
    \begin{itemize}
        \item \textit{Keywords}: placement, misaligned, incorrect angle, orientation
        \item \textit{Example}: "The curtain is hanging in the air instead of the bar."
    \end{itemize}

    \item \textbf{Missing/Extra Objects}: Captures cases where objects are unexpectedly added or removed, causing inconsistencies.  
    \begin{itemize}
        \item \textit{Keywords}: missing, removed, added, extra, inconsistent
        \item \textit{Example}: "The man’s face was removed and replaced by a mask."
    \end{itemize}

    \item \textbf{Edges}: Focuses on issues related to sharp, uneven, or poorly blended edges.  
    \begin{itemize}
        \item \textit{Keywords}: edges, sharp, uneven, jagged
        \item \textit{Example}: "The edges of the pizza are not even."
    \end{itemize}

    \item \textbf{Resolution}: Refers to cases where the visual clarity or quality of the image is degraded, often appearing pixelated or with visual noise.  
    \begin{itemize}
        \item \textit{Keywords}: resolution, clarity, pixelated, low quality
        \item \textit{Example}: "The image of the bird looks pixelated and low in resolution."
    \end{itemize}
\end{itemize}

\subsection{Analysis Methodology}

Our categorization process followed these steps:

\begin{enumerate}
    \item \textbf{Examining the Workers' Feedback}:  
    We reviewed detailed textual feedback from workers who evaluated the instruction-based edits. Each piece of feedback was carefully analyzed to identify recurring issues.

    \item \textbf{Identifying Categories}:  
    We identified common themes in the feedback and organized them into meaningful categories representing distinct visual and technical issues.

    \item \textbf{Extracting Keywords for Categories}:  
    For each category, we identified specific keywords and phrases that workers frequently used to describe the issues. These keywords were used to group similar feedback together.

    \item \textbf{Generating Statistics}:  
    We quantified the frequency of each category across the entire dataset to understand which types of issues were most prevalent. This analysis provided insights to guide future improvements in the edits.
\end{enumerate}

\subsection{Supervision Details}
\label{sec:supervision_details}
The model was fine-tuned for 1 epoch using AdamW with a \(2 \times 10^{-4}\) learning rate. Since it accepts a single image input, we concatenated the before-and-after images.
\subsection{Model Versions}
\label{sec:model_versions}
\begin{itemize}
    \item \textbf{GPT Models}
    \begin{itemize}
        \item GPT-4o (Released: 2024-08-06)
        \item GPT-4 Turbo (Released: 2024-04-09)
        \item GPT-4 (Version: 0613)
    \end{itemize}

    \item \textbf{Gemini Models}
    \begin{itemize}
        \item Gemini 1.5 Pro (001)
        \item Gemini 1.0 Pro (001)
    \end{itemize}

    \item \textbf{Image Editing Models}
    \begin{itemize}
        \item \texttt{imagen-3.0-capability-001} via Vertex AI API, guidance scale set to 12
        \item UltraEdit (Original publicly available version)
    \end{itemize}
\end{itemize}
\subsection{Additional Experiments}
\label{sec:additional_experiments}
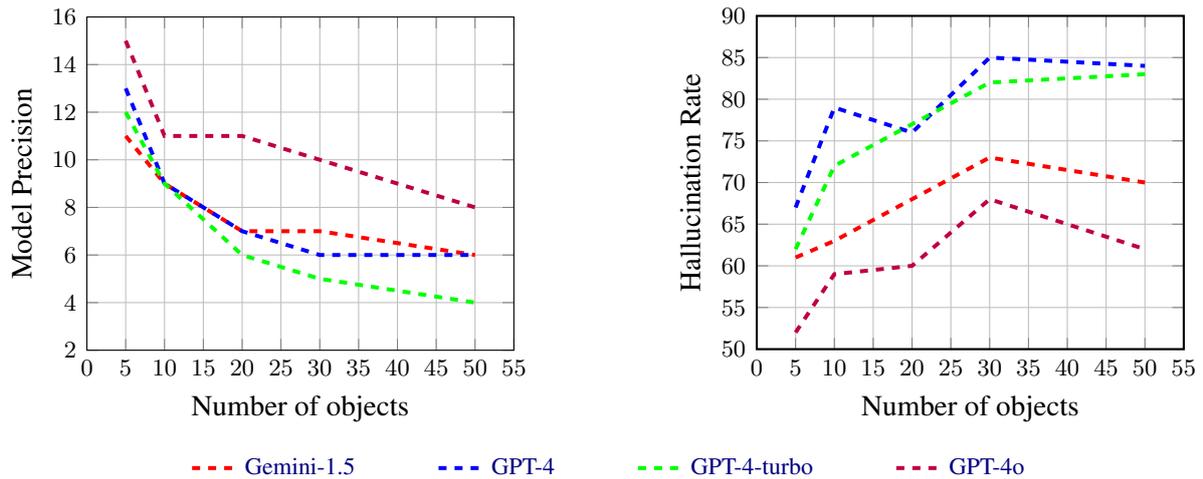
\begin{figure*}[tp]
    \centering
    \begin{minipage}[t]{0.45\textwidth}
    \begin{tikzpicture}
    \begin{axis}[
        xlabel={Number of objects},
        ylabel={Model Precision},
        grid=major,
        width=\textwidth,
        height=6cm,
        mark options={solid},
        xtick={0,5,...,50,55},
        ytick={0,2, 4,...,16},
        ymin=2, ymax=16,
        xmin=0, xmax=55,
        tick label style={font=\footnotesize},
        legend to name=combinedlegend, 
        legend columns=-1, 
        legend style={font=\footnotesize, draw=none, /tikz/every even column/.append style={column sep=1cm}} 
    ]
    
    \addplot[dashed, color=red, line width=1.5pt]  
        coordinates {
        (5, 11.0) (10, 9.0) (20, 7.0) (30, 7.0) (50, 6.0)
        };
    \addlegendentry{Gemini-1.5}

    \addplot[dashed, color=blue, line width=1.5pt]  
        coordinates {
        (5, 13.0) (10, 9.0) (20, 7.0) (30, 6.0) (50, 6.0)
        };
    \addlegendentry{GPT-4}

    \addplot[dashed, color=green, line width=1.5pt]  
        coordinates {
        (5, 12.0) (10, 9.0) (20, 6.0) (30, 5.0) (50, 4.0)
        };
    \addlegendentry{GPT-4-turbo}

    \addplot[dashed, color=purple, line width=1.5pt] 
        coordinates {
        (5, 15.0) (10, 11.0) (20, 11.0) (30, 10.0) (50, 8.0)
        };
    \addlegendentry{GPT-4o}
    
    \end{axis}
\end{tikzpicture}
    \end{minipage}
    \hfill
     \begin{minipage}[t]{0.45\textwidth}
        \begin{tikzpicture}
            \begin{axis}[
                xlabel={Number of objects},
                ylabel={Hallucination Rate},
                grid=major,
                width=\textwidth,
                height=6cm,
                xtick={0,5,...,55},
                ytick={50,55,...,90},
                ymin=50, ymax=90,
                xmin=0, xmax=55,
                tick label style={font=\footnotesize},
                thick,
                legend to name=combinedlegend, 
                legend columns=-1, 
                legend style={font=\footnotesize, draw=none, /tikz/every even column/.append style={column sep=1cm}} 
            ]
            \addplot[dashed, color=red, line width=1.5pt] 
                coordinates {(5, 61.0) (10, 63.0) (20, 68.0) (30, 73.0) (50, 70.0)};
            \addlegendentry{Gemini-1.5}
            \addplot[dashed, color=blue, line width=1.5pt] 
                coordinates {(5, 67.0) (10, 79.0) (20, 76.0) (30, 85.0) (50, 84.0)};
            \addlegendentry{GPT-4}
            \addplot[dashed, color=green, line width=1.5pt] 
                coordinates {(5, 62.0) (10, 72.0) (20, 77.0) (30, 82.0) (50, 83.0)};
            \addlegendentry{GPT-4-turbo}
            \addplot[dashed, color=purple, line width=1.5pt] 
                coordinates {(5, 52.0) (10, 59.0) (20, 60.0) (30, 68.0) (50, 62.0)};
            \addlegendentry{GPT-4o}
            \end{axis}
        \end{tikzpicture}
    \end{minipage}
    
    \begin{center}
        \ref{combinedlegend}
    \end{center}
    
    \caption{Comparison of model precision and hallucination rates as a function of the number of objects in the edited images. The performance of all models decreases as the number of objects in the images increases, highlighting a trend where more complex scenes contribute to higher hallucination rates and lower precision.}
    \label{fig:number_of_objects_comparison}
\end{figure*}
\sparagraph{0.6ex}{Hallucination Rates as a Function of the Number of Objects}

Figure~\ref{fig:number_of_objects_comparison} presents the precision and hallucination rates as a function of the number of objects in the edited
images. There is a performance drop in all models as the number of objects in the images increases, highlighting a trend where more complex scenes contribute to higher hallucination rates and lower precision.

\sparagraph{0.6ex}{Out-of-Distribution Evaluation of the EditInspector Model}

To evaluate the out-of-distribution (OOD) performance of our fine-tuned model, we conducted an additional experiment using a balanced set of 120 samples from the Image Editing Request (IER) dataset. The IER dataset consists of human-authored edits, originally collected from Reddit, and created using tools like Photoshop. Each edit is paired with a free-form language instruction, covering a wide range of real-world edits across both local and global scenarios.

We selected 30 high-quality, localized edits that are content-safe and appropriate for public release. Each edit was then reversed by swapping the source and target images and inverting the instruction, resulting in 60 examples. This approach allowed us to both increase the number of test cases and evaluate the model’s ability to handle edit directionality and instruction symmetry.

For each edit (original and reversed), we also created a manually crafted distractor instruction and difference caption, introducing plausible but incorrect variations. For example, if the original instruction removed an elephant, the distractor might reference removing a nearby person. This setup challenges the model to distinguish accurate vs. misleading edits in realistic and diverse scenarios.

We tested both the base LLaVA model and our fine-tuned model on two key tasks:  
(1) \textbf{Edit Accuracy}, and  
(2) \textbf{Difference Caption Accuracy}.

\textbf{Edit Accuracy Task:}  
The fine-tuned model achieved a balanced accuracy of 59.2\%, compared to 54.2\% for the base model.

Notably, the base model showed a strong bias toward answering "Yes" on nearly all examples (111 out of 120), resulting in a true negative rate of only 3.3\%. In contrast, the fine-tuned model exhibited a more balanced response, predicting "Yes" for 63 out of 120 edits, with a significantly higher true negative rate of 56.6\%.

\textbf{Difference Caption Accuracy Task:}  
The base model again defaulted to "Yes" for all edits, yielding a 0\% true negative rate.

The fine-tuned model provided more calibrated responses, splitting answers approximately evenly and achieving a true negative rate of 56.7\%.

These results demonstrate that our fine-tuned model generalizes better to unfamiliar editing distributions. It shows a stronger grasp of task semantics, provides more accurate judgments in OOD scenarios, and avoids the base model’s tendency to over-predict positive responses.

\subsection{Augmentation methods}
\label{sec:appendix_auggmentation_methods}
\begin{figure*}[h]
    \centering
    \includegraphics[width=1\textwidth]{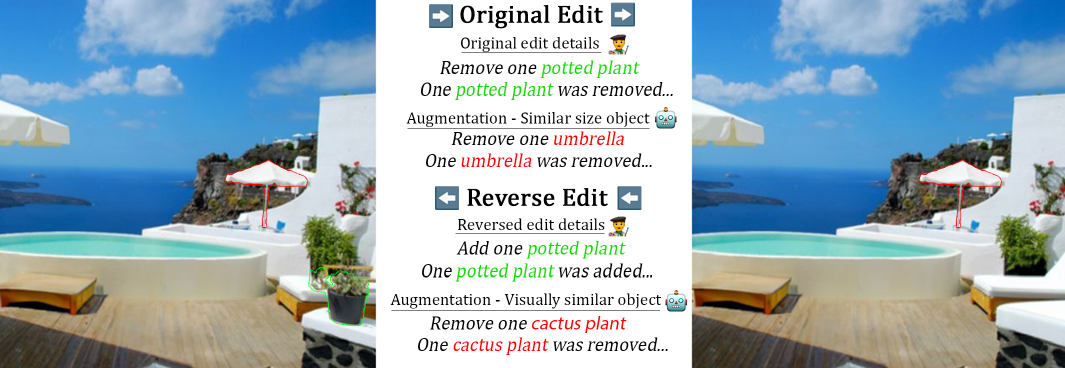}
    \caption{Illustration of our augmentation methods for a remove edit. The pre-edit image (left) shows a potted plant, while the post-edit image (right) depicts the scene with the plant removed. In the first augmentation method, the instruction and difference caption is modified by replacing the ``potted plant'' with an object of similar size (umbrella). In the second augmentation, we reverse the edit by switching the order of the images, changing the instruction and difference caption from ``remove potted plant'' to ``add potted plant,'' and introducing a negative instruction for a visually similar object (e.g., cactus plant), which is absent in the post-edit image.}
    \label{fig:augmentation_example}
\end{figure*}

\subsection{Licenses}
\label{sec:licenses}
All use of scientific artifacts is consistent with their intended use. This work focuses on evaluating existing models in the English language using images from the MagicBrush dataset and does not introduce new models, generate new images, or employ technologies that could pose ethical, societal, or safety risks. We collected anonymous human annotations using Amazon Mechanical Turk crowdsourcing platform. The images are used in accordance with the MagicBrush license, and the evaluation code and dataset are released under the CC-BY-4.0 license.
\subsection{Annotation UI}
\label{sec:appendx_annotation_ui}
\begin{figure*}[bp]
\begin{center}
\includegraphics[width=\textwidth,height=8cm, keepaspectratio]{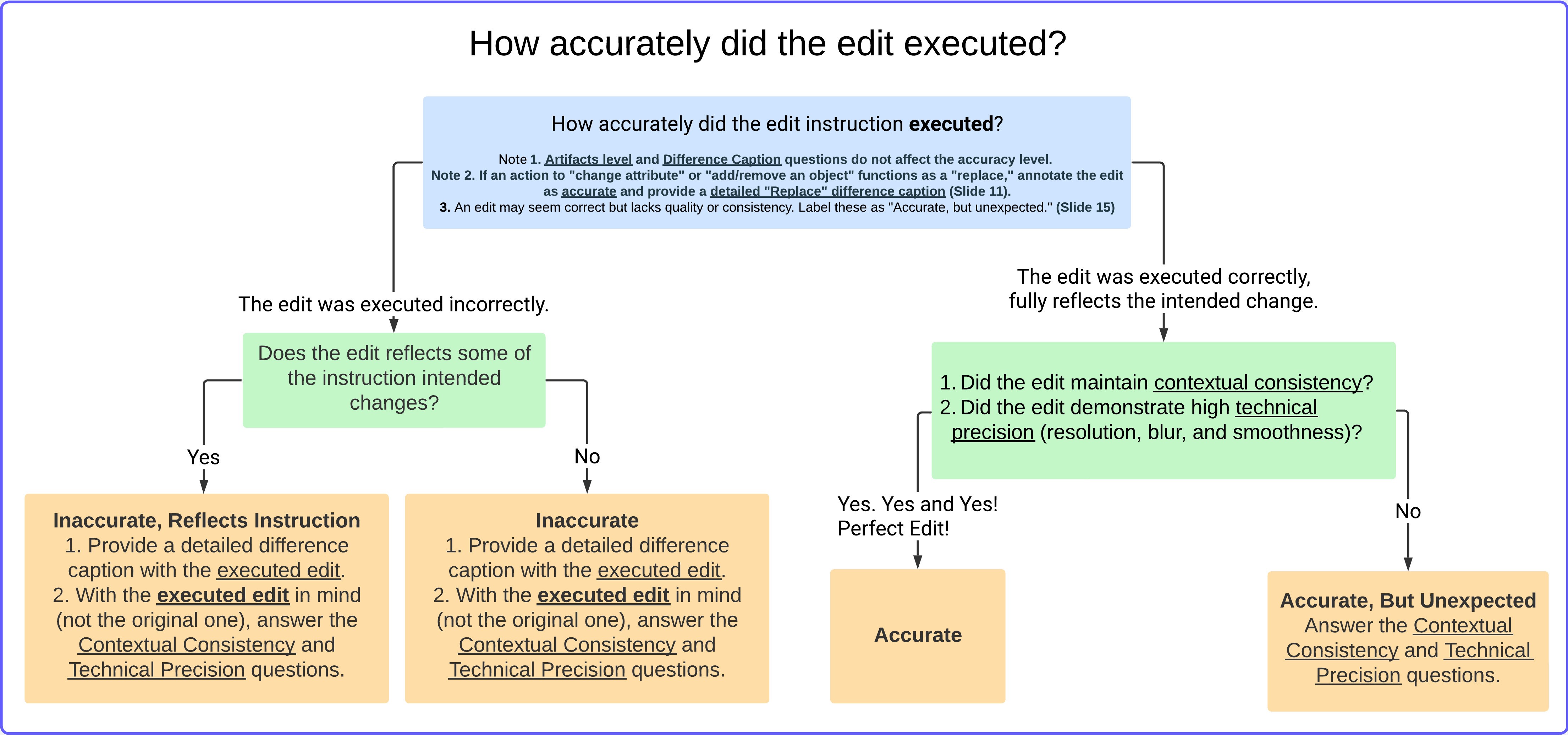}
\end{center}
\caption{The accuracy scheme tree that was provided to annotators to guide the answering process.}
\label{fig:accuracy-tree}
\end{figure*}

\begin{figure*}[!h]
\begin{center}
\includegraphics[width=\textwidth,height=8cm, keepaspectratio]{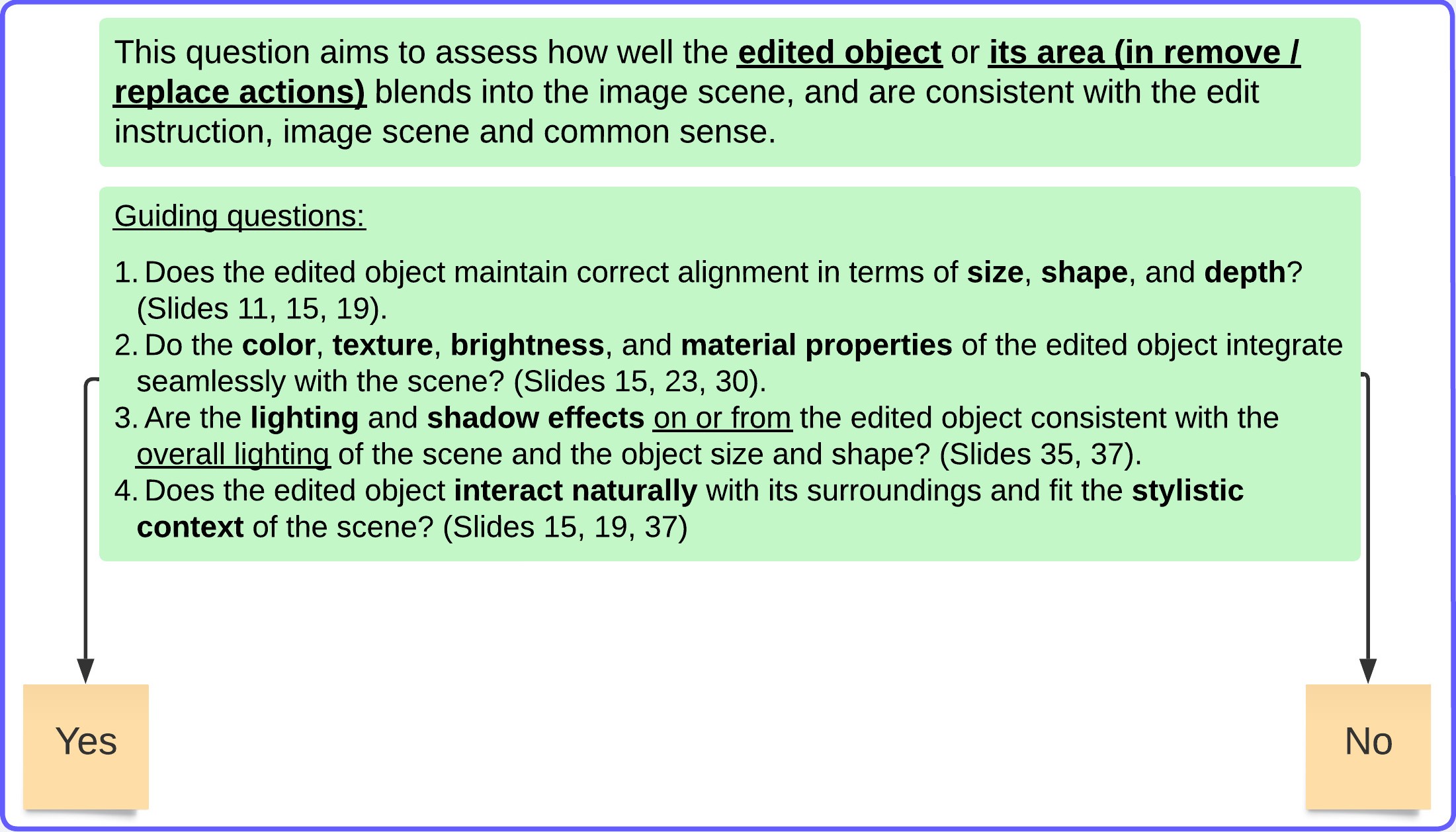}
\end{center}
\caption{The contextual consistency scheme tree that was provided to annotators to guide the answering process.}
\label{fig:contextual-consisteny-tree}
\end{figure*}

\begin{figure*}[!h]
\begin{center}
\includegraphics[width=\textwidth,height=8cm, keepaspectratio]{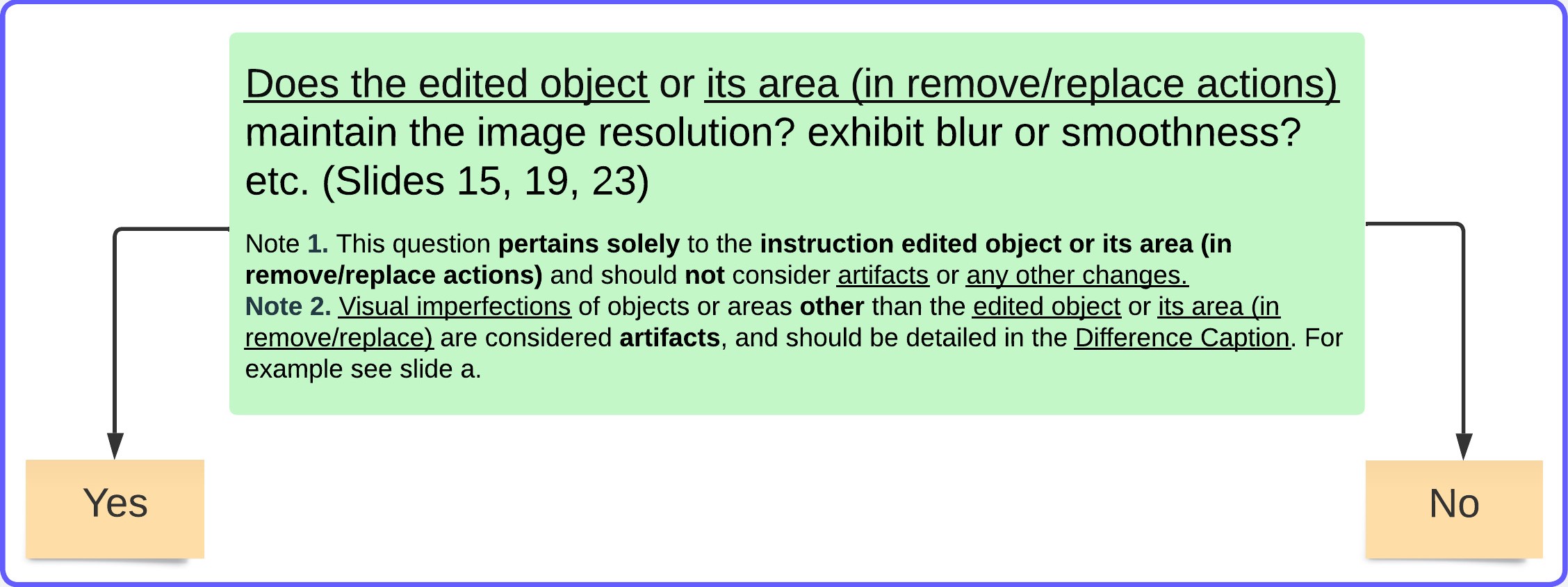}
\end{center}
\caption{The technical precision scheme tree that was provided to annotators to guide the answering process.}
\label{fig:technical-precision-tree}
\end{figure*}

\begin{figure*}[h]
\begin{center}
\includegraphics[width=\textwidth,height=8cm, keepaspectratio]{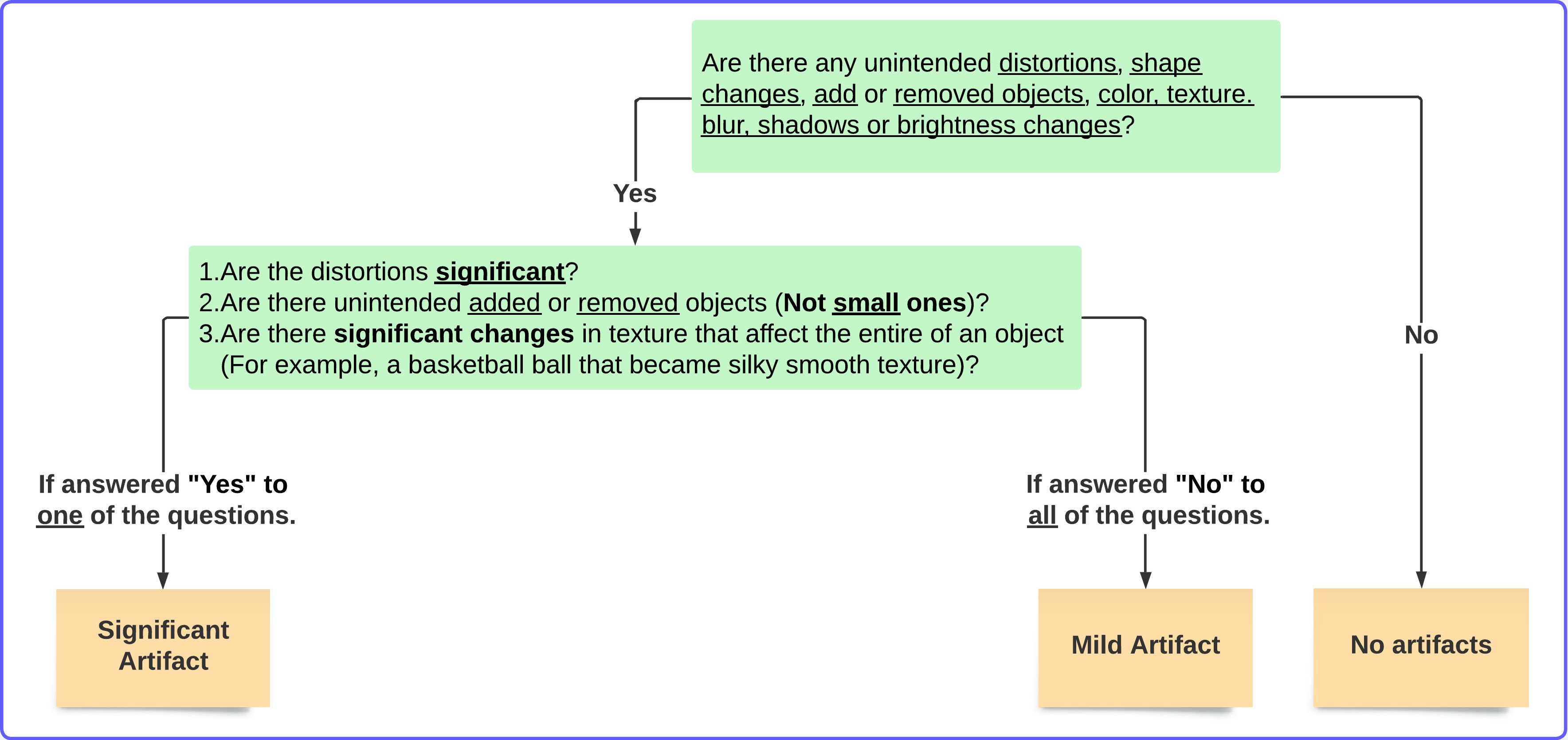}
\end{center}
\caption{The artifacts scheme tree that was provided to annotators to guide the answering process.}
\label{fig:artifcats-tree}
\end{figure*}

\begin{figure*}[h]
\begin{center}
\includegraphics[width=\textwidth,height=8cm, keepaspectratio]{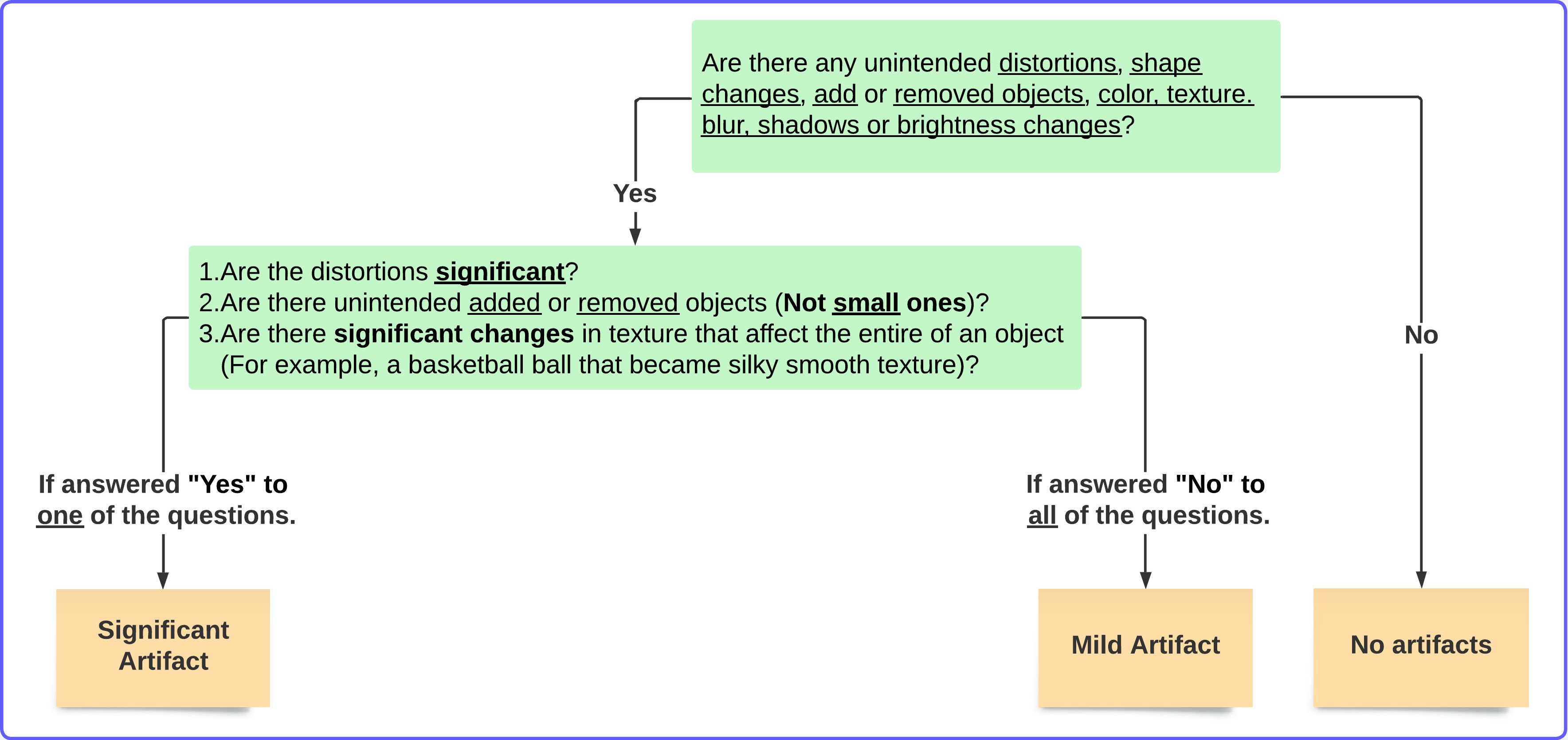}
\end{center}
\caption{The difference caption instructions provided to annotators to guide the answering process.}
\label{fig:difference-caption-instruction-window}
\end{figure*}

\begin{figure*}[h]
\begin{center}
\includegraphics[width=\textwidth,height=8cm, keepaspectratio]{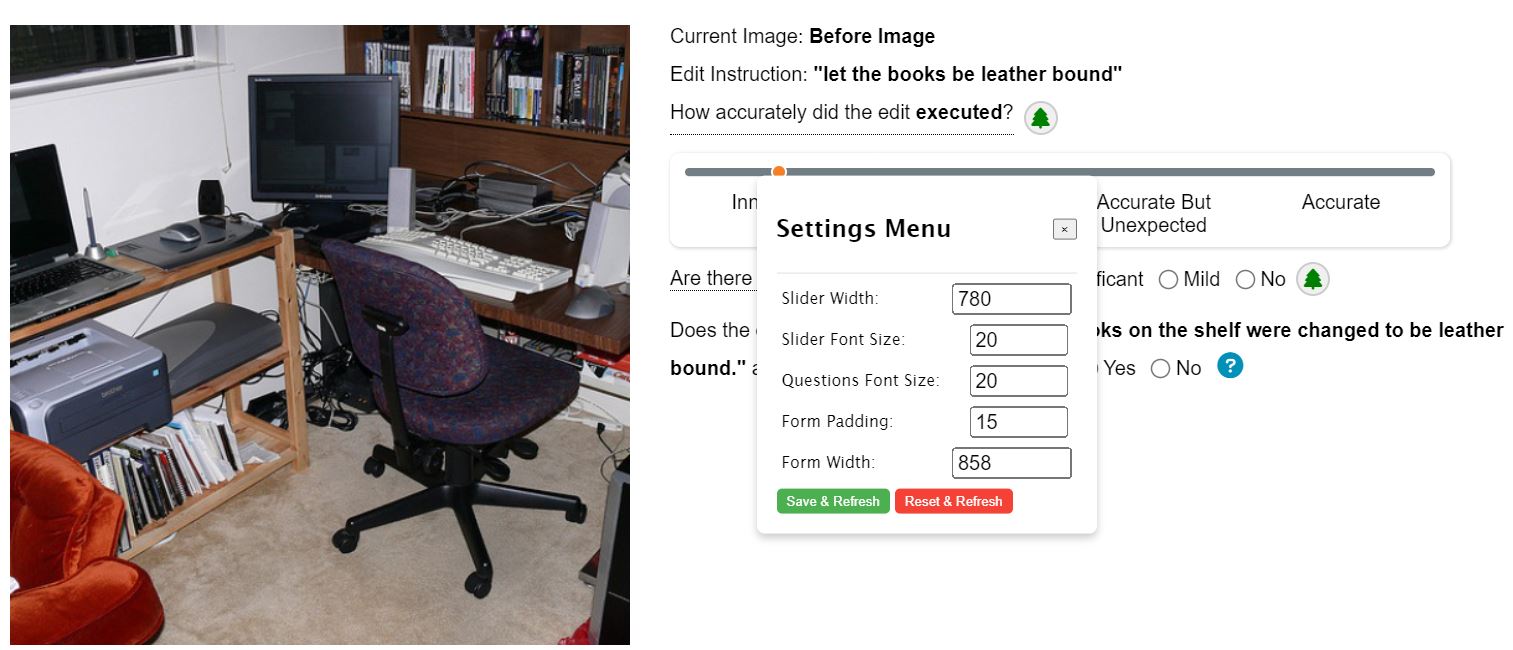}
\end{center}
\caption{The setting menu for customizing the form font size, width etc.}
\label{fig:setting-menu-window}
\end{figure*}
\subsection{Annotation Examples}
\label{sec:appendx_annotation_examples}
\begin{figure*}[h!]
\begin{center}
\includegraphics[width=\textwidth,height=8cm, keepaspectratio]{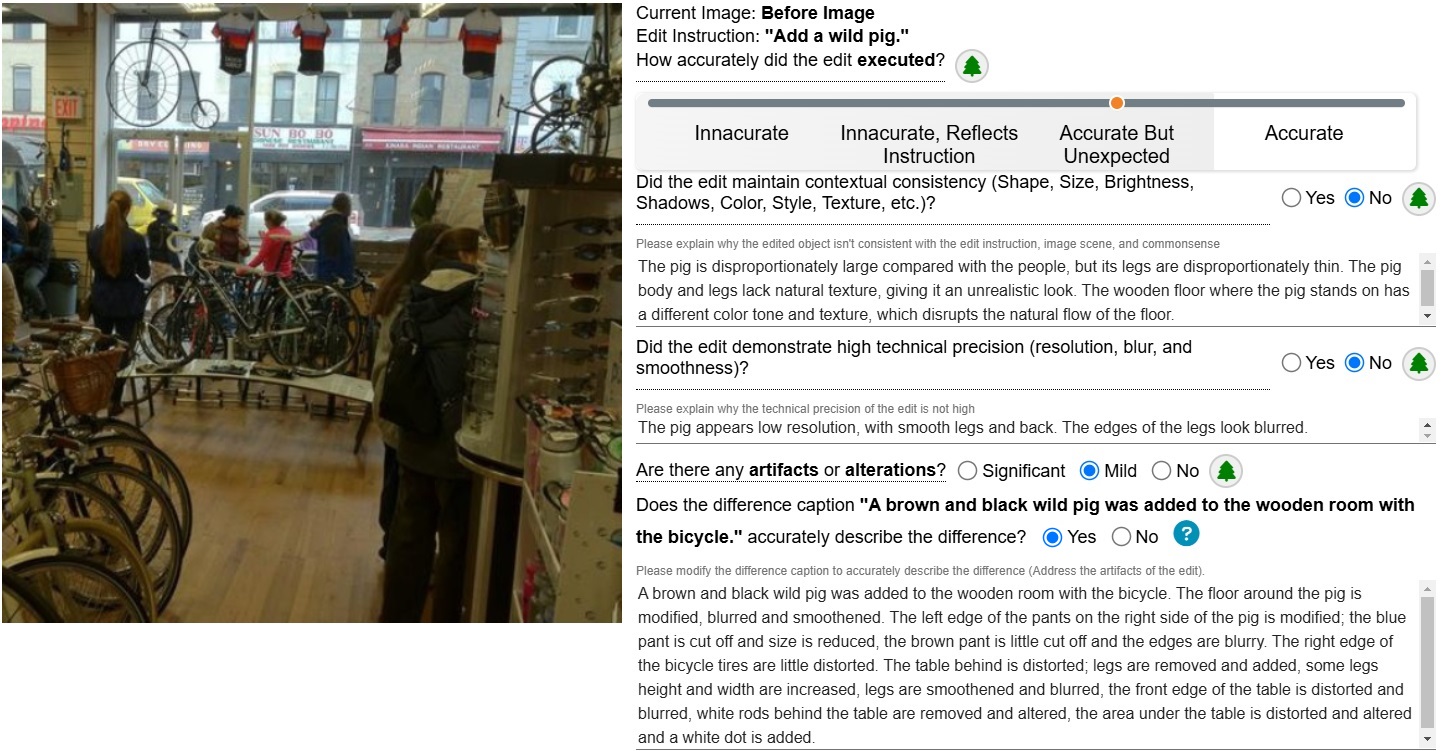}
\end{center}
\caption{Example of image edit verification sample - before image (Add a wild pig).}
\label{fig:add-wild-pig-before}
\end{figure*}

\begin{figure*}[h!]
\begin{center}
\includegraphics[width=\textwidth,height=8cm, keepaspectratio]{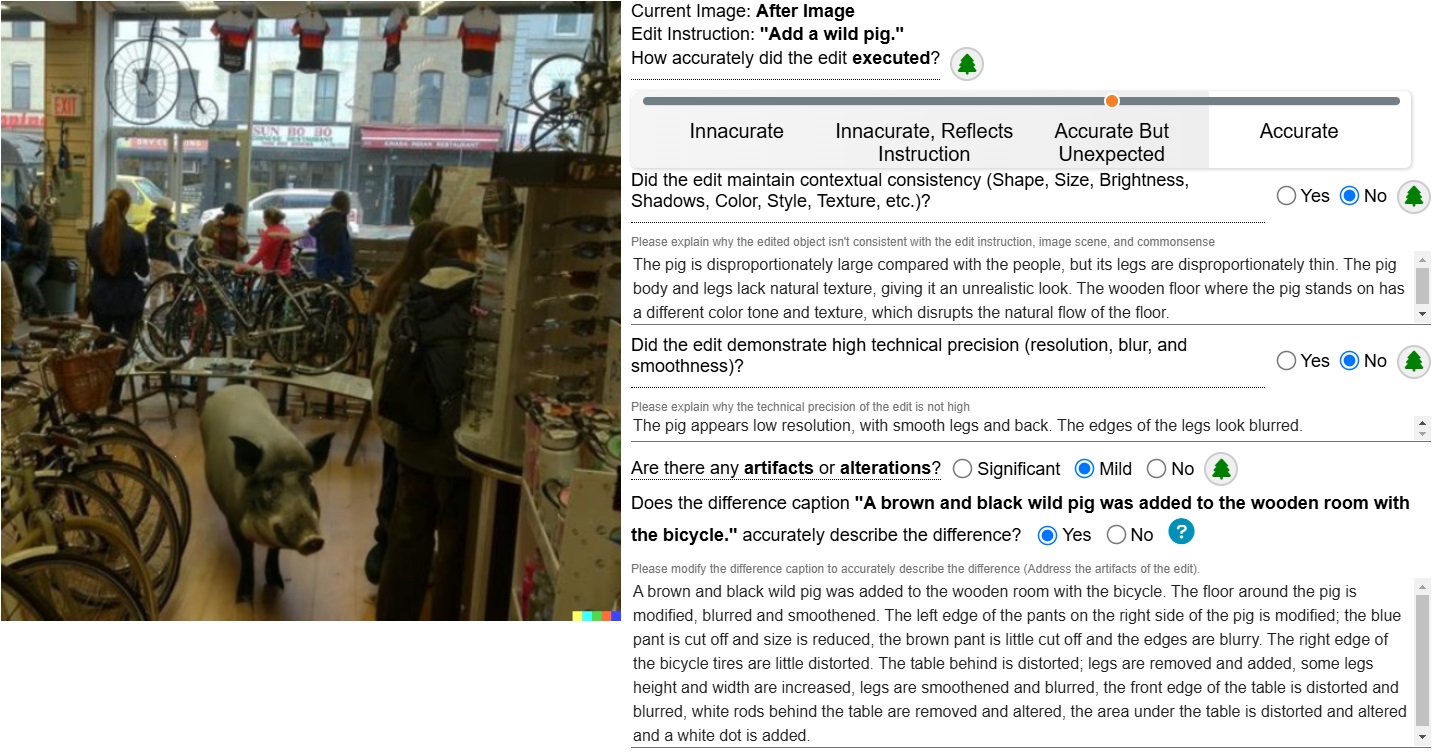}
\end{center}
\caption{Example of image edit verification sample - after image (Add a wild pig).}
\label{fig:add-wild-pig-after}
\end{figure*}

\begin{figure*}[h!]
\begin{center}
\includegraphics[width=\textwidth,height=8cm, keepaspectratio]{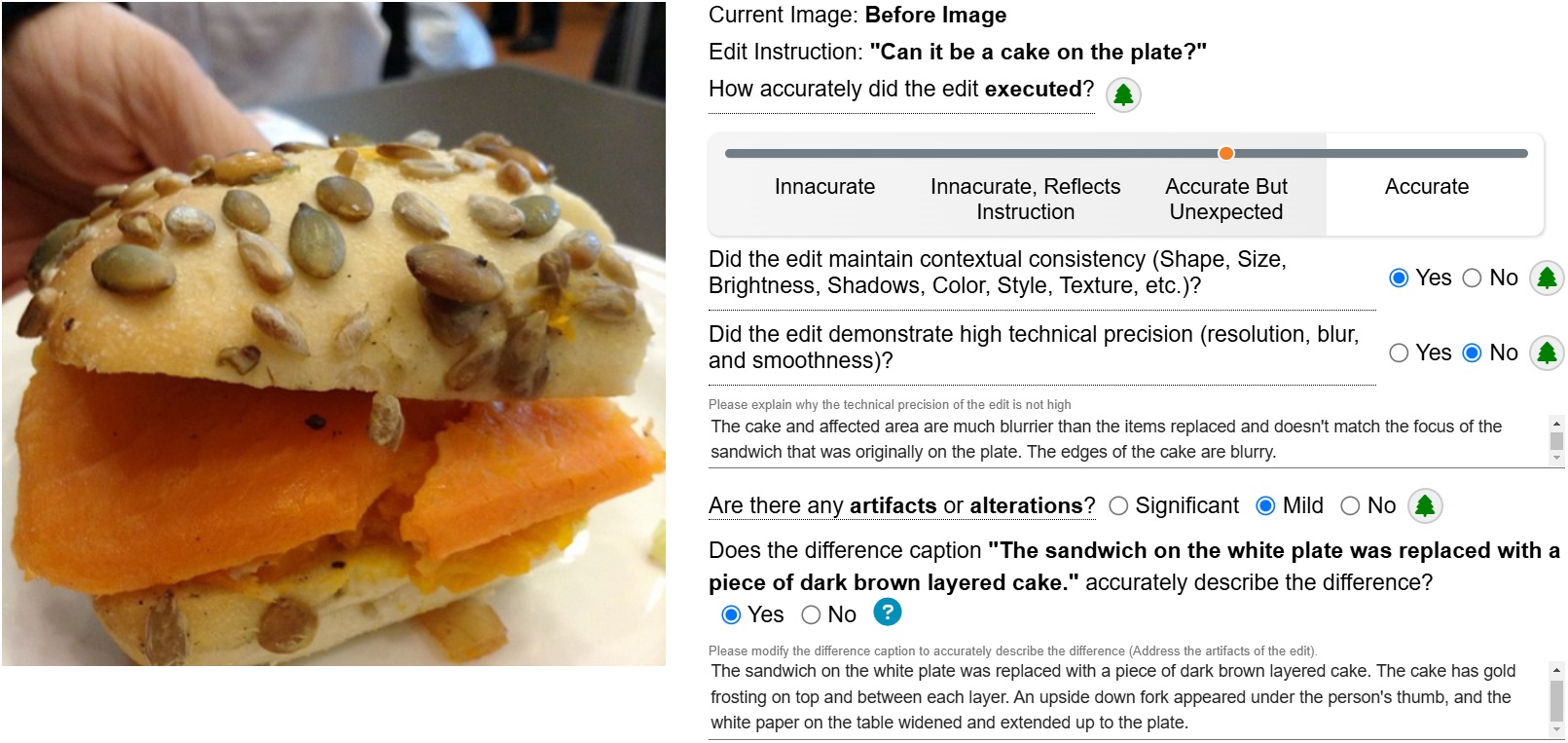}
\end{center}
\caption{Example of image edit verification sample - before image (Cake on the plate).}
\label{fig:cake-plate-before}
\end{figure*}

\begin{figure*}[h!]
\begin{center}
\includegraphics[width=\textwidth,height=8cm, keepaspectratio]{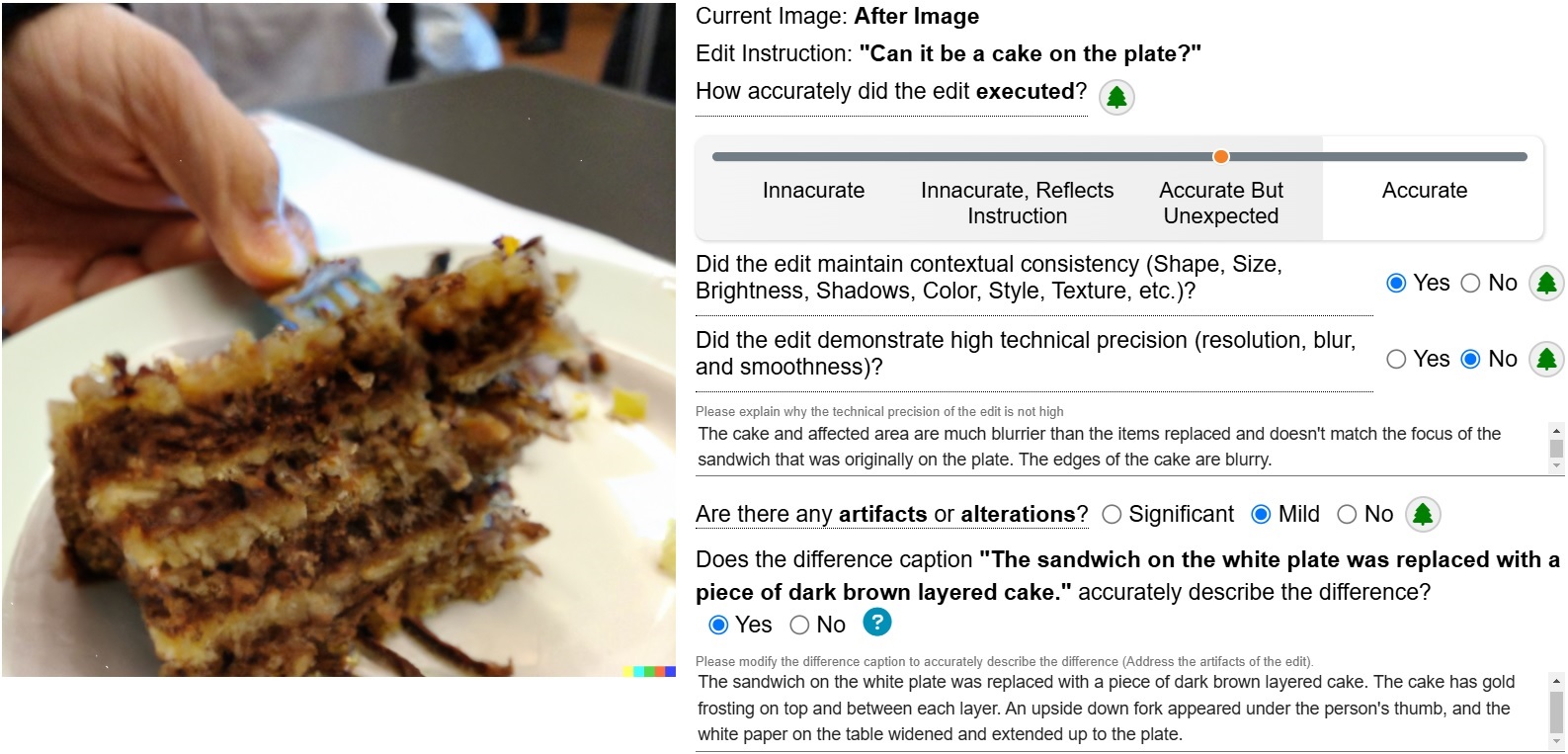}
\end{center}
\caption{Example of image edit verification sample - after image (Cake on the plate).}
\label{fig:cake-plate-after}
\end{figure*}

\begin{figure*}[h!]
\begin{center}
\includegraphics[width=\textwidth,height=8cm, keepaspectratio]{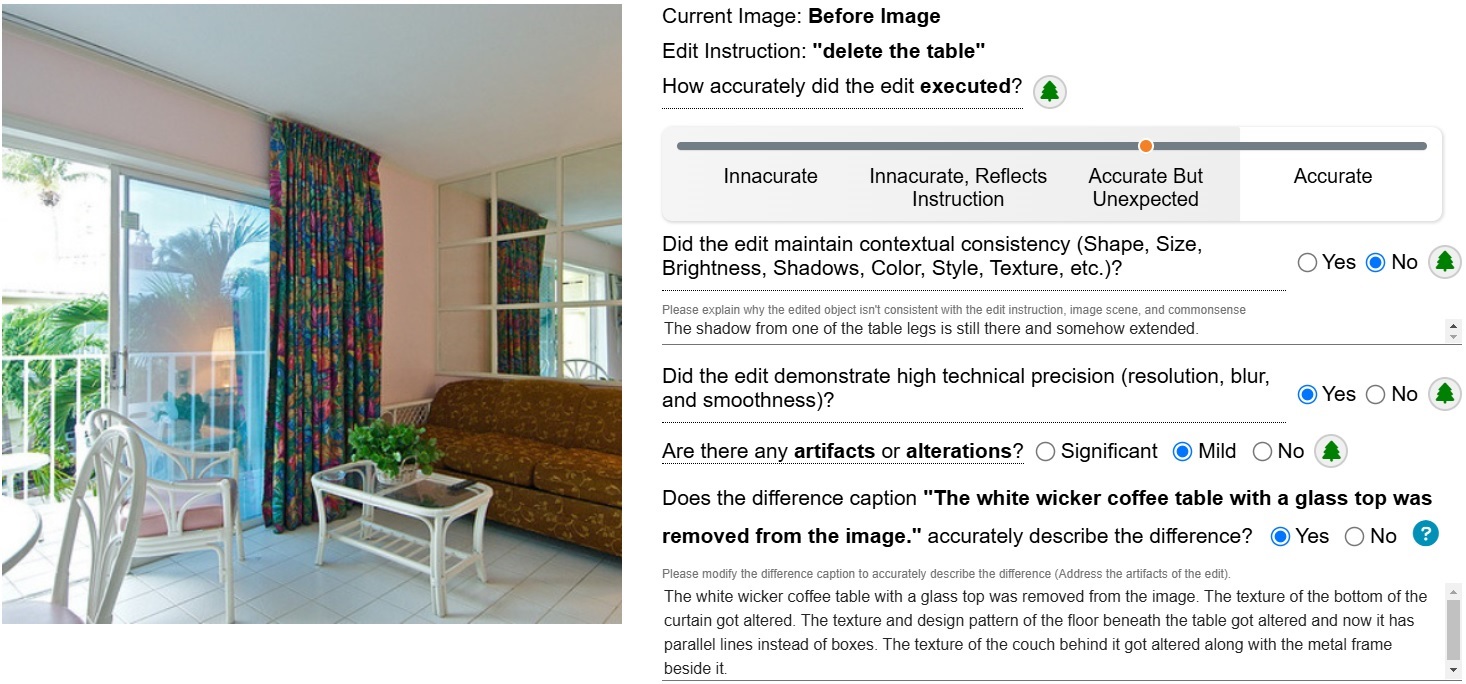}
\end{center}
\caption{Example of image edit verification sample - before image (Delete the table).}
\label{fig:delete-table-before}
\end{figure*}

\begin{figure*}[h!]
\begin{center}
\includegraphics[width=\textwidth,height=8cm, keepaspectratio]{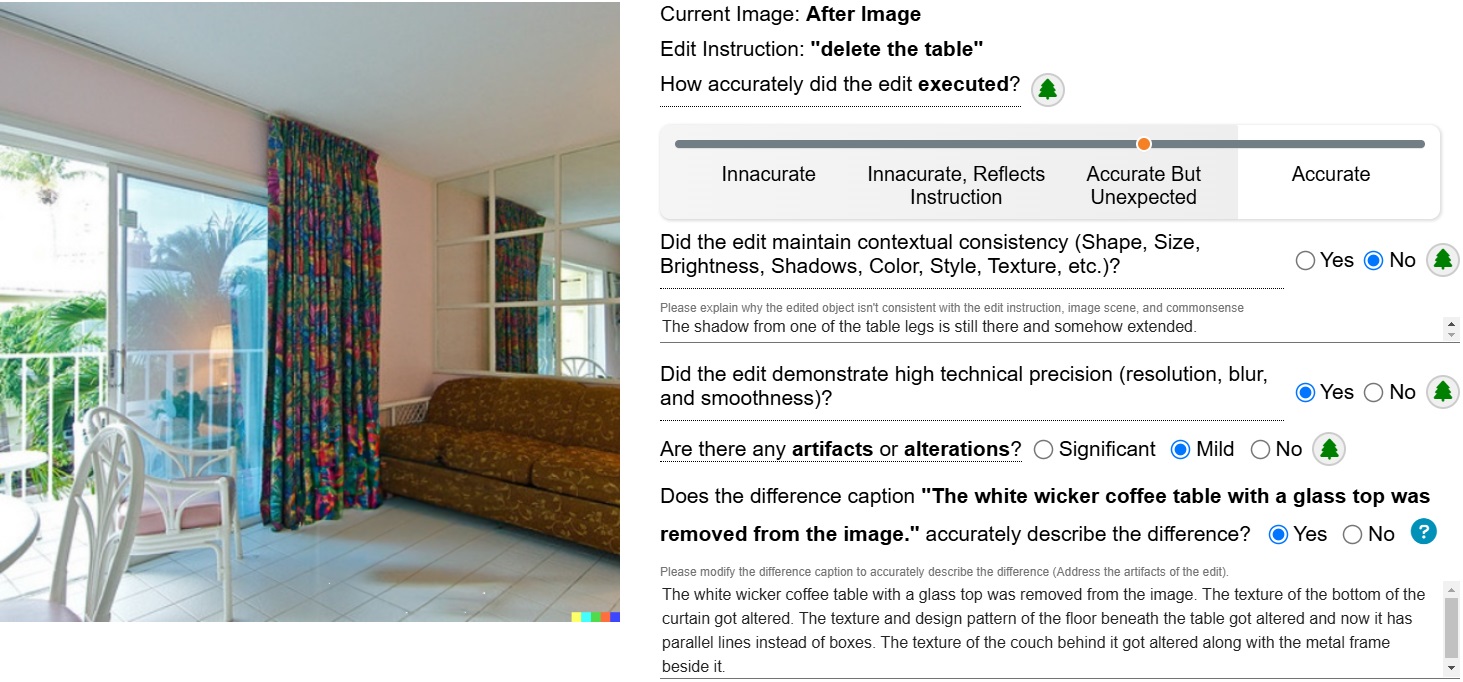}
\end{center}
\caption{Example of image edit verification sample - after image (Delete the table).}
\label{fig:delete-table-after}
\end{figure*}

\begin{figure*}[h!]
\begin{center}
\includegraphics[width=\textwidth,height=8cm, keepaspectratio]{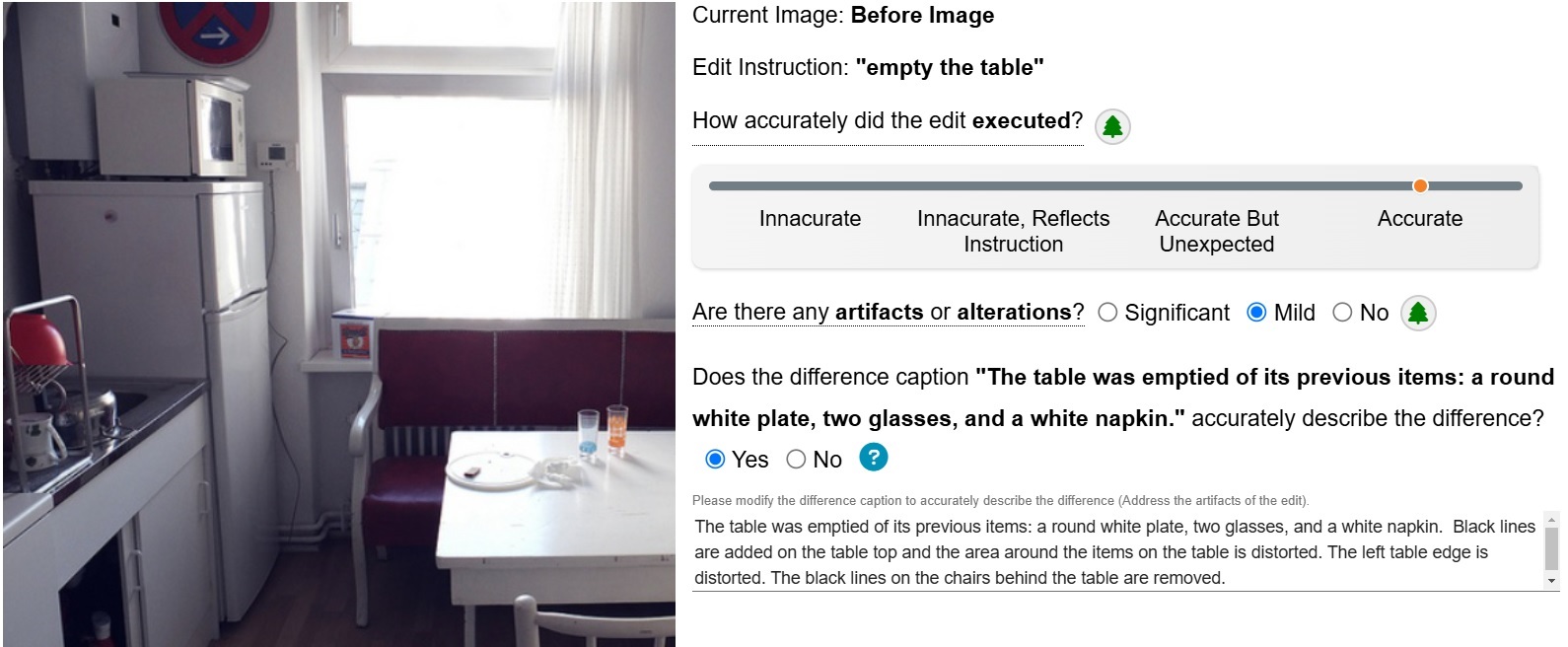}
\end{center}
\caption{Example of image edit verification sample - before image (Empty the table).}
\label{fig:empty-table-before}
\end{figure*}

\begin{figure*}[h!]
\begin{center}
\includegraphics[width=\textwidth,height=8cm, keepaspectratio]{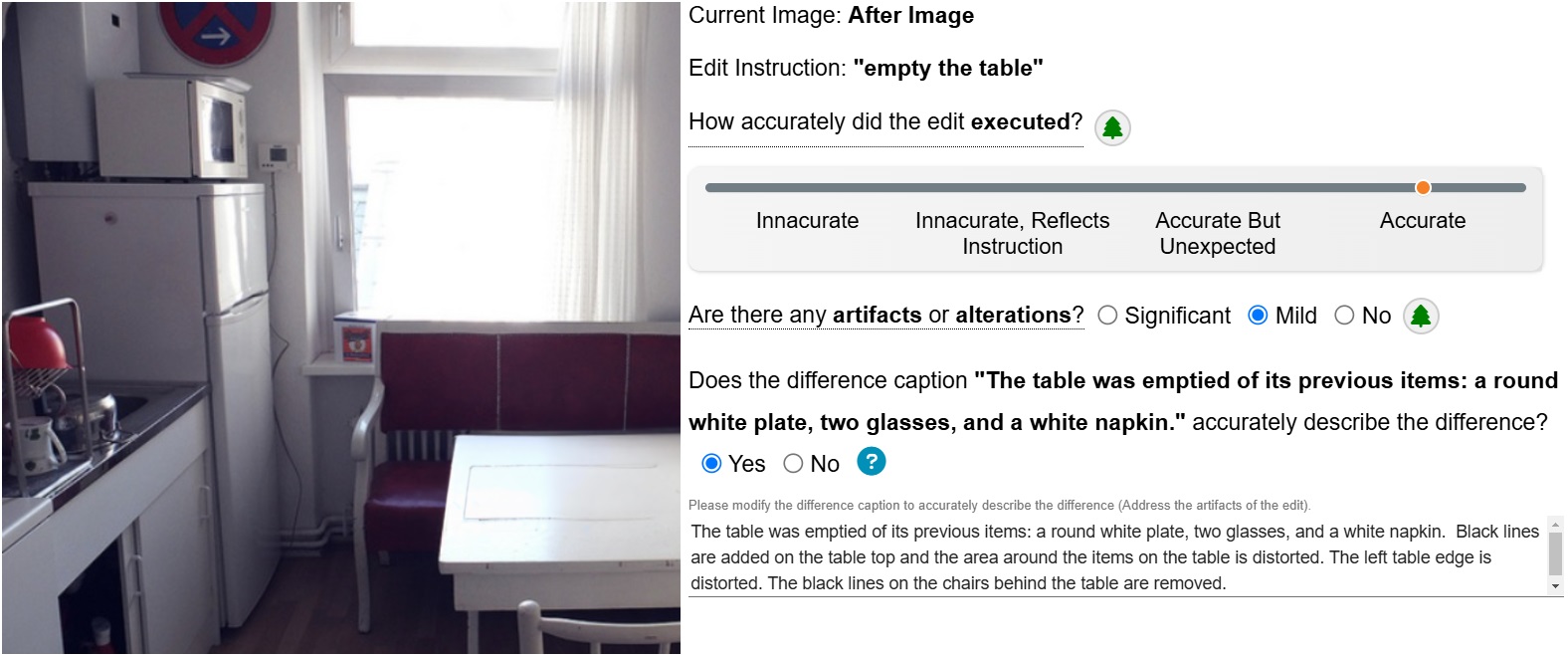}
\end{center}
\caption{Example of image edit verification sample - after image (Empty the table).}
\label{fig:empty-table-after}
\end{figure*}

\begin{figure*}[h!]
\begin{center}
\includegraphics[width=\textwidth,height=8cm, keepaspectratio]{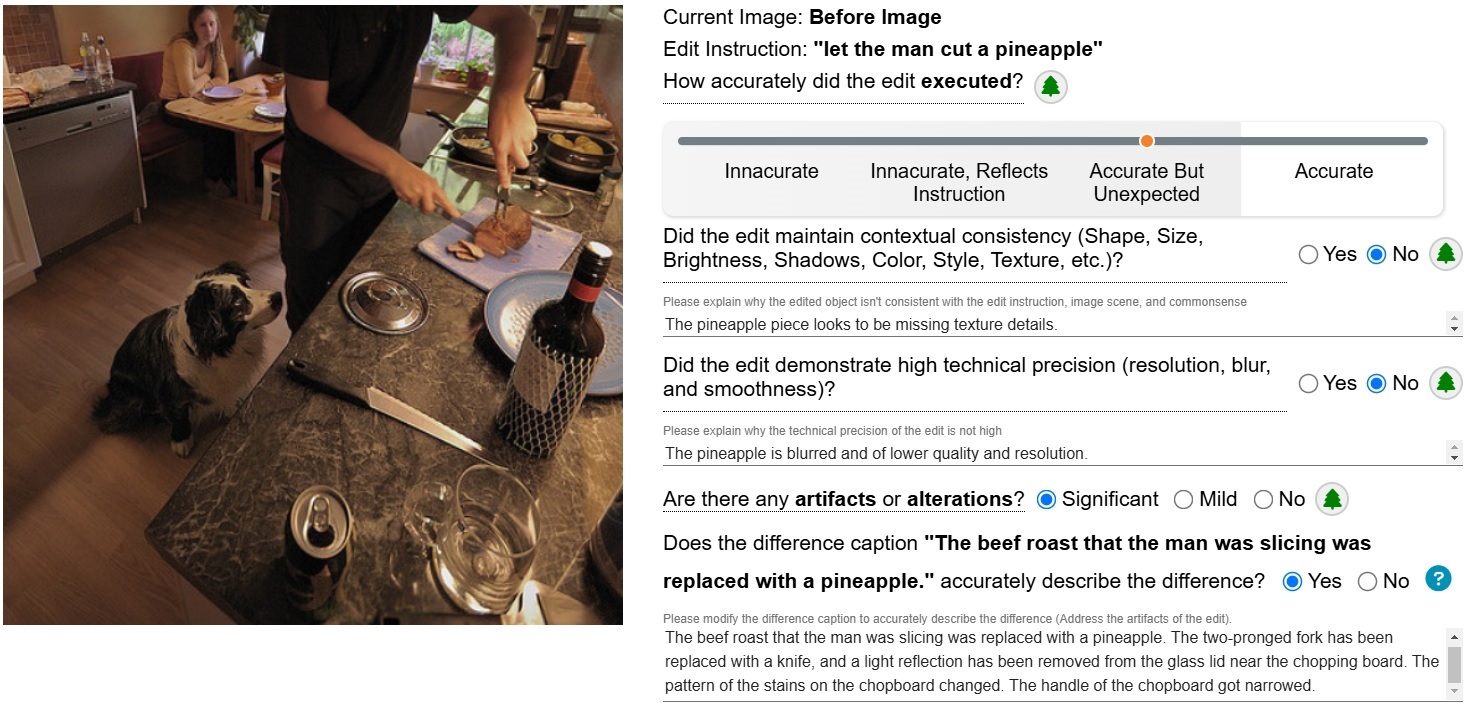}
\end{center}
\caption{Example of image edit verification sample - before image (Cut a pineapple).}
\label{fig:cut-pineapple-before}
\end{figure*}

\begin{figure*}[h!]
\begin{center}
\includegraphics[width=\textwidth,height=8cm, keepaspectratio]{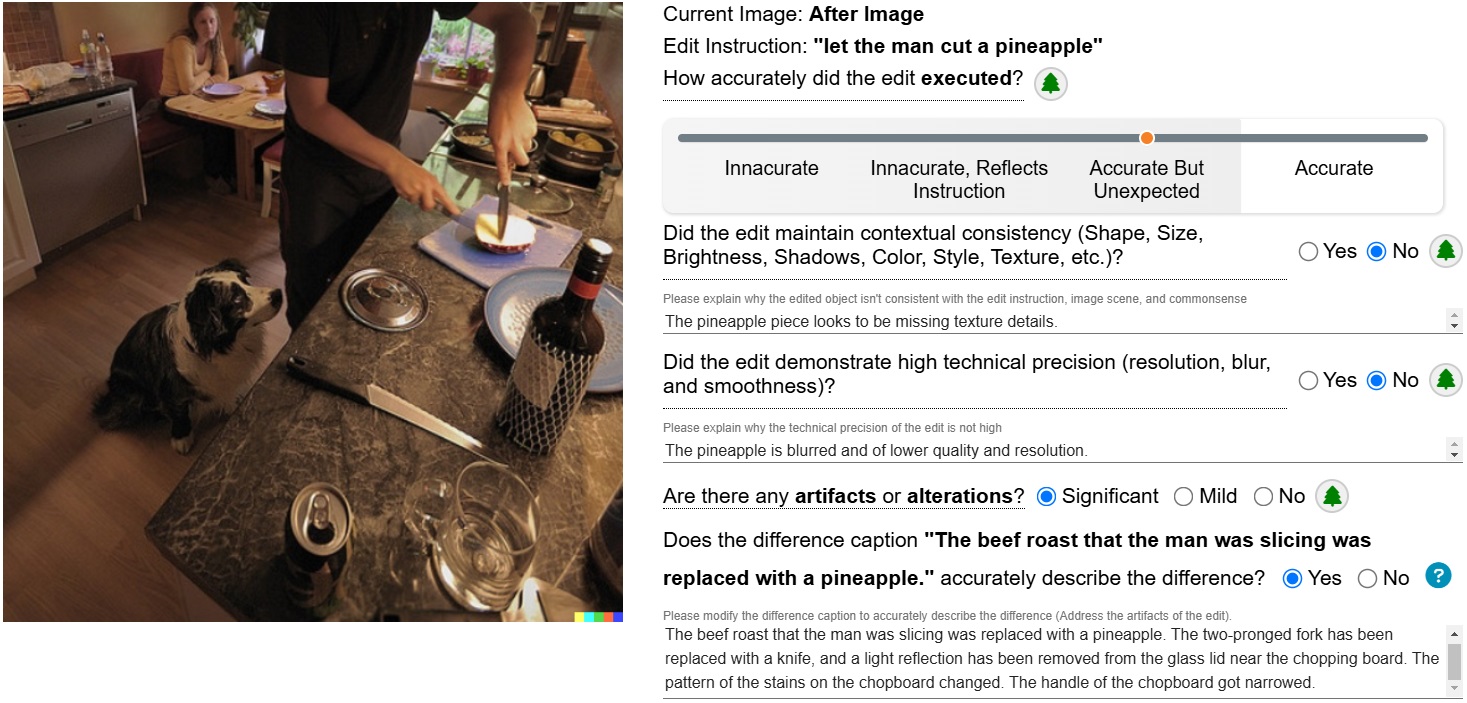}
\end{center}
\caption{Example of image edit verification sample - after image (Cut a pineapple).}
\label{fig:cut-pineapple-after}
\end{figure*}

\end{document}